\documentclass[10pt,twocolumn,letterpaper]{article}
\usepackage[pagenumbers]{cvpr}

\usepackage[T1]{fontenc}
\usepackage[utf8]{inputenc}
\usepackage[final]{microtype}

\usepackage{amsmath,amssymb,amsfonts,amsthm,mathtools}
\usepackage{graphicx}
\usepackage{multirow}
\usepackage{makecell}
\usepackage{array}
\usepackage{float}
\usepackage{placeins}
\usepackage{dblfloatfix}
\usepackage[normalem]{ulem}
\usepackage{algorithm}
\usepackage{algpseudocode}
\usepackage{stmaryrd}
\usepackage{tikz}
\usepackage{enumitem}
\usepackage{bm}
\usepackage{dsfont} 
\usepackage{stmaryrd}
\usepackage{cuted}
\usepackage{caption}
\newcommand{\sd}[1]{\,{\scriptsize$\pm$#1}}
\newcommand{\pp}[1]{\,{\scriptsize(#1)}}

\usetikzlibrary{
  arrows.meta,
  calc,
  backgrounds,
  positioning,
  decorations.pathreplacing
}

\DeclareMathSizes{5}{5}{3.5}{3}

\definecolor{surfblue}{RGB}{20,86,196}
\definecolor{spherefill}{RGB}{120,168,240}
\definecolor{hitgreen}{RGB}{22,148,58}
\definecolor{validorange}{RGB}{214,118,24}
\definecolor{invalidred}{RGB}{194,48,48}
\definecolor{patchviolet}{RGB}{126,47,142}
\definecolor{epsviolet}{RGB}{126,47,142}
\definecolor{sensordk}{RGB}{24,24,24}
\definecolor{raygray}{RGB}{112,112,112}
\definecolor{refc}{RGB}{0,114,178}
\definecolor{srcc}{RGB}{213,94,0}
\definecolor{surfc}{RGB}{120,120,120}
\definecolor{planec}{RGB}{0,158,115}
\definecolor{xc}{RGB}{190,30,45}      

\newtheoremstyle{compact}
  {4pt}
  {4pt}
  {\itshape}
  {}
  {\bfseries}
  {.}
  {0.5em}
  {}

\theoremstyle{compact}

\newcommand{\Lip}{\operatorname{Lip}}

\usepackage[table]{xcolor}
\usepackage{booktabs}
\usepackage{graphicx}

\definecolor{bestcolor}{RGB}{255,205,205}     
\definecolor{secondcolor}{RGB}{255,226,180}   
\definecolor{thirdcolor}{RGB}{255,244,190}    

\newcommand{\best}[1]{\cellcolor{bestcolor}#1}
\newcommand{\second}[1]{\cellcolor{secondcolor}#1}
\newcommand{\third}[1]{\cellcolor{thirdcolor}#1}

\def\confName{CVPR}
\def\confYear{2027}

\definecolor{cvprblue}{rgb}{0.21,0.49,0.74}
\usepackage[pagebackref,breaklinks,colorlinks,allcolors=cvprblue]{hyperref}
\usepackage{xr-hyper}
\title{First-Hit Photoconsistency for Neural Surface Reconstruction}

\author{
Gabriel Lenain\\
Inria Paris\\
École normale supérieure -- PSL\\
{\tt\small gabriel.lenain@inria.fr}
\and
Edmond Boyer\\
Univ. Grenoble Alpes, Inria\\
CNRS, Grenoble INP, LJK\\
{\tt\small edmond.boyer@inria.fr}
\and
Jean Ponce\\
École normale supérieure -- PSL\\
New York University\\
{\tt\small jean.ponce@ens.fr}
}

\date{}

\begin{document}
\maketitle
\begin{strip}
    \centering
    \includegraphics[width=\textwidth]{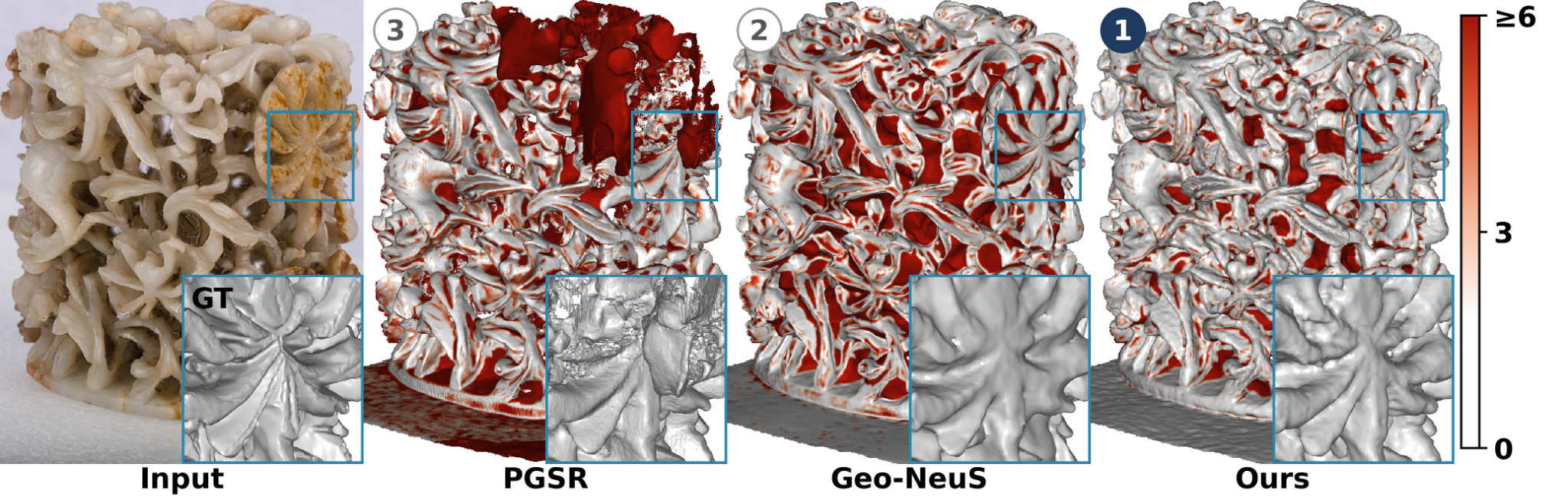}
    \vspace{-6mm}
    \captionof{figure}{Surface reconstruction on the BMVS Jade scene. We compare our method with PGSR, based on Gaussian splatting and Geo-NeuS based on neural implicit volumetric rendering. Colors indicate the distance from the reconstructed surfaces to the reference mesh ($\times 10^3$, normalized scene units), clipped at the colorbar maximum. Circled
numbers rank methods by Chamfer distance.}
    \label{fig:teaser}
\end{strip}

\begin{figure*}[t]
    \centering
    \includegraphics[width=\textwidth]{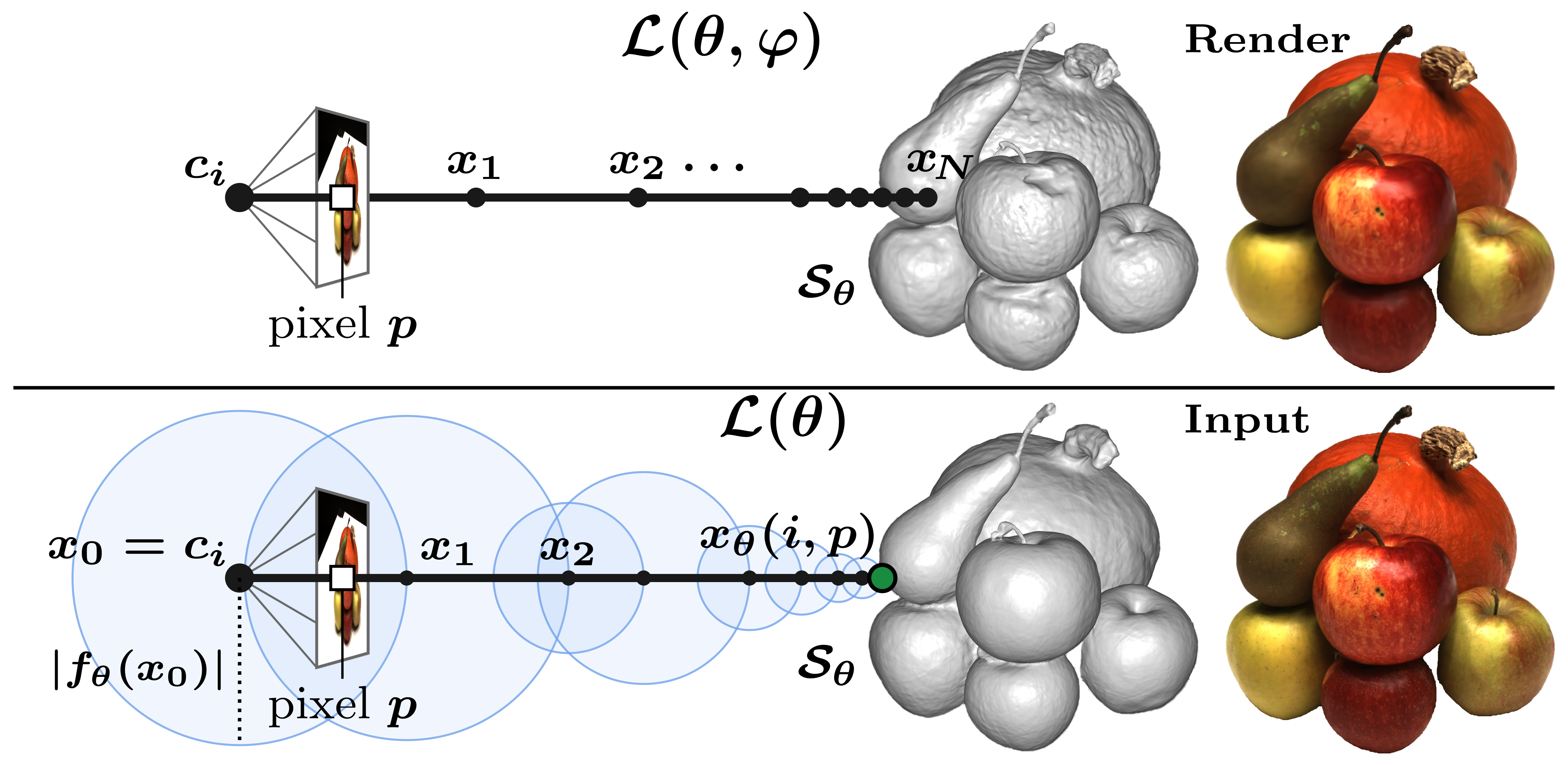}
    \vspace{-6mm}
    \caption{
Top: Volumetric rendering jointly optimizes geometry $\theta$ and appearance $\varphi$
by compositing colors at hierarchically sampled points $x_1, \dots, x_N$ along each ray (black dots).
Bottom: Our method directly optimizes the first surface intersections
$x_\theta(i,p)$, found by sphere tracing (blue spheres), through patch photoconsistency.
On DTU scan 63, NeuS (top), a representative SDF-based method,
produces a rendering (Render) visually similar to the observed
camera image (Input), despite geometric inaccuracies.
}
    \label{fig:vol_vs_ours}
\end{figure*}

\begin{abstract}
\noindent 
Recent neural implicit and Gaussian splatting approaches have achieved impressive results in multi-view surface reconstruction. However, jointly optimizing geometry and appearance introduces ambiguities. A parametric appearance model can compensate for geometric errors, reproducing the input images accurately without an accurate surface. We demonstrate accurate neural implicit surface reconstruction using only multi-view patch photoconsistency with geometry-based visibility, without volumetric rendering or parametric appearance models. For each reference ray, sphere tracing computes the first surface intersection. Its position and normal define a local patch whose photoconsistency is evaluated across visible neighboring views using colors sampled directly from the input images. For efficient sphere tracing, we use a 1-Lipschitz multi-layer perceptron. We use a single photoconsistency loss, without foreground masks or auxiliary geometric regularization. Experiments show that our method achieves the lowest mean Chamfer distance on BlendedMVS (BMVS) among compared methods and matches the geometric accuracy of leading reconstruction methods on DTU.
\end{abstract}

\section{Introduction}
Reconstructing a surface from multiple calibrated images is a fundamental problem in 3D vision. Multi-view stereo (MVS) has traditionally recovered geometry through image matching and geometric consistency ~\cite{furukawa2010pmvs,schoenberger2016pixelwise}. Neural implicit representations have subsequently become a standard tool, providing continuous, resolution-independent representations of geometry. Early methods optimized signed distance or occupancy fields through differentiable surface rendering~\cite{yariv2020idr,niemeyer2020dvr}. Later approaches, including NeuS~\cite{wang2021neus} and VolSDF~\cite{yariv2021volsdf}, adopted volumetric rendering, converting the implicit representation into a density and accumulating colors along camera rays.  More recently, 3D Gaussian splatting~\cite{kerbl2023gaussian} has enabled efficient rendering using explicit Gaussian primitives. Geometry-oriented variants, including 2DGS~\cite{huang2024_2dgs}, PGSR~\cite{chen2025pgsr} and AmbiSuR~\cite{li2026ambisur} aim to improve surface reconstruction while retaining opacity-based compositing along camera rays.
\\
In these methods, photometric supervision is applied to rendered images rather than directly to the reconstructed surface, potentially allowing geometric inaccuracies to persist (Fig.~\ref{fig:vol_vs_ours}, top). Moreover, parametric appearance models, such as color MLPs or spherical harmonics, can compensate for geometric errors by adjusting the predicted appearance. Methods such as NeuralWarp \cite{darmon2022neuralwarp} and Geo-NeuS \cite{fu2022geoneus} address these limitations by introducing explicit multi-view patch photoconsistency alongside volumetric rendering, improving geometric accuracy. This raises the question of whether patch photoconsistency alone can provide sufficient supervision to optimize a continuous neural surface, without volumetric rendering or a parametric appearance model.
\\
To investigate this question, we directly optimize a continuous neural surface through multi-view patch photoconsistency, building on the principle of photometric agreement between corresponding image regions~\cite{faugeras1998complete,vu2012highaccuracy,furukawa2010pmvs}. Specifically, we represent the surface as the zero level set of a 1-Lipschitz MLP and use sphere tracing~\cite{hart1996sphere} to compute the first surface intersection along each camera ray (Fig.~\ref{fig:vol_vs_ours}, bottom). At each intersection, the surface position and normal define a local tangent-plane patch whose photoconsistency is evaluated across neighboring views, with visibility explicitly determined from the evolving surface geometry. Colors are sampled directly from the input images through differentiable interpolation, without a parametric appearance network. Since patch photoconsistency relies on sufficiently accurate initial correspondences, existing methods start from a sparse set of matched keypoints~\cite{furukawa2010pmvs} or refine an existing reconstruction~\cite{vu2012highaccuracy,darmon2022neuralwarp}. Similarly, we initialize the implicit field from a coarse geometric proxy obtained by classical MVS from the same input images, then refine its zero level set using only multi-view patch photoconsistency. 
\\
Our method achieves the lowest mean Chamfer distance on BMVS and matches the geometric accuracy of leading reconstruction methods on DTU. 
\\ 
The main contributions of this paper are:
\begin{enumerate}
    \item A geometry-only optimization formulation of neural surface reconstruction based solely on multi-view patch photoconsistency, without
volumetric rendering, a parametric appearance model or
auxiliary geometric regularization.

    \item A differentiable first-hit optimization framework that explicitly evaluates source-view visibility against the evolving neural surface.

\end{enumerate}

\section{Related Work}

\par\smallskip\noindent\textbf{Multi-view stereo and variational refinement.}
Classical MVS recovers geometry by enforcing
photometric consistency across calibrated views together with explicit
visibility reasoning, whether by
expanding and filtering patches from matched seeds~\cite{furukawa2010pmvs}
or estimating per-view depth maps and fusing geometrically consistent
estimates~\cite{merrell2007realtime,galliani2015massively,schoenberger2016pixelwise,xu2022acmmp}.
Variational approaches optimize surfaces directly through multi-view
photoconsistency, evolving level sets on volumetric
grids~\cite{faugeras1998complete,pons2007multi}
or refining triangular meshes through photometric reprojection
error~\cite{vu2012highaccuracy,delaunoy2014photometric},
typically with smoothness regularization.
Like these methods, we optimize the surface directly through multi-view photoconsistency, but represent it as a continuous neural field rather than a grid or mesh, explicitly determine visibility using first-hit intersections and require no auxiliary geometric regularization.

\par\smallskip\noindent\textbf{Neural surface reconstruction.}
Neural implicit methods represent geometry as the level set of a signed distance or occupancy field and optimize it by comparing rendered colors with observed images, typically using a parametric appearance model. Colors are evaluated either at the first surface intersection~\cite{yariv2020idr,niemeyer2020dvr} or, following NeRF~\cite{mildenhall2020nerf}, integrated along camera rays~\cite{wang2021neus,yariv2021volsdf}. Later
work improves geometric detail and
representation capacity~\cite{li2023neuralangelo}.
To address the geometric ambiguity of photometric rendering, several methods introduce additional constraints. MVSDF incorporates stereo-derived depth and feature consistency~\cite{zhang2021mvsdf}, while NeuralWarp and Geo-NeuS combine volumetric rendering with patch photoconsistency~\cite{darmon2022neuralwarp,fu2022geoneus}. Geometry-oriented Gaussian splatting methods~\cite{huang2024_2dgs,chen2025pgsr,li2026ambisur} similarly add geometric constraints to opacity-based rendering.

\par\smallskip\noindent\textbf{Differentiable surface rendering.}
Optimizing implicit surfaces through ray intersections requires both differentiating the intersection points and computing them reliably. Differentiable surface rendering methods propagate gradients through these intersections using implicit differentiation~\citep{yariv2020idr,niemeyer2020dvr} or approximate sphere-tracing gradients~\citep{liu2020dist}. Other approaches address visibility discontinuities through reparameterization~\cite{vicini2022differentiable,bangaru2022differentiable}. Sphere tracing~\cite{hart1996sphere} finds the first surface intersection along a ray using steps bounded by the distance to the surface. Eikonal regularization encourages SDF behavior~\cite{gropp2020igr} but does not guarantee such bounds, and its optimization can exhibit instabilities~\cite{yang2023steik}.

\par\smallskip\noindent\textbf{Lipschitz neural networks.}
A 1-Lipschitz field guarantees that sphere-tracing steps
do not exceed the distance to the surface. Expressive
1-Lipschitz architectures include GroupSort~\cite{anil2019groupsort},
which uses norm-preserving activations, and convex
potential layers (CPL)~\cite{meunier2022cpl,prach2024layers}.
Lenain et al.~\citep{lenain2026surface} fit 1-Lipschitz
neural fields to range scans using first-hit depth regression
and Coiffier and Béthune~\citep{coiffier2024lipschitz}
fit such fields to point clouds or triangle soups.
While we also use a 1-Lipschitz architecture, we address
surface reconstruction from calibrated RGB images through
multi-view patch photoconsistency with explicit
cross-view visibility.

\section{Method}

\subsection{Problem statement}

Given $n$ calibrated images $\{I_i:\Omega\to\mathbb{R}^3_+\}_{i=1}^n$, where
$\Omega\subset\mathbb{R}^2$ is the image domain, we estimate the surface $S$
as the zero level set $S_\theta=\{x\in\mathbb{R}^3 : f_\theta(x)=0\}$ of a
neural field $f_\theta:\mathbb{R}^3\to\mathbb{R}$ with parameters
$\theta\in\mathbb{R}^d$. We take $f_\theta$ to be negative inside the object
and positive outside.

\paragraph{Notation.}
Let $c_i \in \mathbb{R}^3$ denote the center of camera $i$,
$u_i(p) \in \mathbb{R}^3$ the unit viewing direction through
pixel $p \in \mathbb{R}^2$ and $\pi_i(x) \in \mathbb{R}^2$ the
projection of a point $x \in \mathbb{R}^3$ into image $i$.
We define the first-hit depth for camera $i$ and pixel $p$ as
\begin{equation}
t_\theta(i,p)
=
\inf\{t\in[0,T]:
f_\theta(c_i+t u_i(p))=0\},
\end{equation}
with maximum tracing distance $T >0$ and $t_\theta(i,p)=\infty$ when no intersection exists.
For a valid intersection, the corresponding surface point is $x_\theta(i,p)=c_i+t_\theta(i,p)u_i(p).$

\subsection{Enforcing photoconsistency}
\label{sec:photo-supervision}

\begin{figure}[!ht]
    \centering
    \includegraphics[width=\columnwidth]{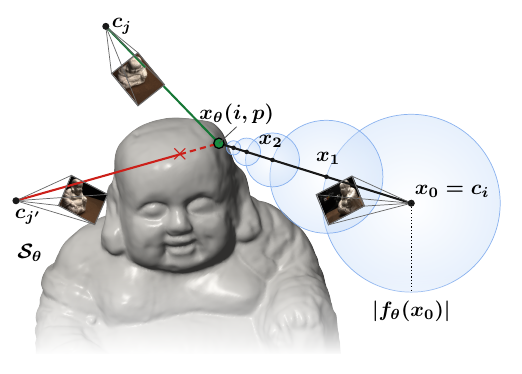}
    \vspace{-6mm}
    \caption{ The first surface intersection $x_\theta(i,p)$ obtained by sphere tracing is visible from $c_j$ (green) but occluded from $c_{j'}$ (red).}
    \label{fig:method}
\end{figure}

We supervise the implicit surface \(f_\theta\) using multi-view patch
photoconsistency. At each optimization step, we uniformly sample reference views
\(i\), trace surface intersections from sampled reference pixels and compare
the patches with those observed in neighboring source views \(j\).
Intensities are evaluated at projected locations by bilinear
interpolation, following differentiable image sampling~\cite{jaderberg2015stn}.
Since photometric comparisons are meaningful only when \(x\) is visible, we define \(\mathcal{V}_\theta(x;i)\) as the set of source views in which a point \(x\), observed from reference view \(i\), is also visible (Fig.~\ref{fig:method}).

\par\smallskip\noindent\textbf{Explicit visibility.}
Let \(x=x_\theta(i,p)\) be the first surface intersection along
the reference ray \((i,p)\). We retain source views \(j\ne i\) for which \(x\) projects inside the image, passes the occlusion test and is viewed at a non-grazing angle.
\\
Formally, the set of valid source views is
\begin{equation}
\label{eq:visible-source-set}
\mathcal{V}_\theta(x;i)
=
\left\{
j \neq i
\;\middle|\;
\begin{aligned}
&\pi_j(x)\in\Omega,\\
&\|x-c_j\|\leq t_\theta(j,\pi_j(x))+\Delta,\\
&n_\theta(x)^\top v_j(x)>\kappa
\end{aligned}
\right\}.
\end{equation}
\noindent
Here
\(n_\theta(x)=\nabla f_\theta(x)/\|\nabla f_\theta(x)\|_2\)
is the outward unit normal,
\(v_j(x)=(c_j-x)/\|c_j-x\|_2\) points toward camera \(j\), 
and the thresholds \(\Delta>0\) and \(\kappa>0\)
absorb the tracing tolerance of visibility queries and reject grazing views, respectively. Source views for which sphere tracing fails to converge
are excluded. To simplify notation, we  write \(\mathcal{V}_i(x)\) instead of \(\mathcal{V}_\theta(x;i)\), leaving the dependence on \(\theta\) implicit.

\par
\begin{figure}[!ht]
\centering
\resizebox{\linewidth}{!}{%
\begin{tikzpicture}[
    scale=0.60,
    >={Stealth[length=1.6mm,width=1.2mm]},
    font=\footnotesize,
    vec/.style={->,line width=0.7pt},
    ray/.style={line width=0.35pt,opacity=0.5},
    cray/.style={line width=0.6pt},
    cam/.style={line width=0.5pt,line join=round},
    smp/.style={circle,inner sep=0pt,minimum size=2.2pt},
    zoom/.style={densely dotted,line width=0.45pt},
  ]

\coordinate (X)  at (0,0);
\coordinate (E1) at (1.00,-0.18);
\coordinate (E2) at (0.50, 0.36);
\coordinate (N)  at (-0.10,1.00);
\def\h{0.42}
\def\q{2}
\def\pl{2.7}

\coordinate (Ci) at (-3.9,4.0);
\coordinate (Cj) at ( 4.2,4.0);

\coordinate (P00) at ($(X)-\q*\h*(E1)-\q*\h*(E2)$);
\coordinate (P10) at ($(X)+\q*\h*(E1)-\q*\h*(E2)$);
\coordinate (P11) at ($(X)+\q*\h*(E1)+\q*\h*(E2)$);
\coordinate (P01) at ($(X)-\q*\h*(E1)+\q*\h*(E2)$);

\filldraw[surfc!60,fill=surfc!12,line width=0.4pt]
  (-3.0,-0.55)
  .. controls (-1.5,-0.10) and (1.2,-0.60) .. (2.9,-0.95)
  -- (3.8,0.45)
  .. controls (2.3,0.90) and (-0.2,1.10) .. (-1.8,0.90)
  -- cycle;
\node[surfc!90!black] at (2.75,-0.62) {$\mathcal S_\theta$};

\filldraw[planec!55!black,fill=planec!12,line width=0.4pt,fill opacity=0.85]
  ($(X)-\pl*\h*(E1)-\pl*\h*(E2)$) --
  ($(X)+\pl*\h*(E1)-\pl*\h*(E2)$) --
  ($(X)+\pl*\h*(E1)+\pl*\h*(E2)$) --
  ($(X)-\pl*\h*(E1)+\pl*\h*(E2)$) -- cycle;

\draw[cray,refc] (Ci) -- (X);
\draw[cray,srcc] (Cj) -- (X);
\draw[ray,refc] (Ci) -- (P00);
\draw[ray,refc] (Ci) -- (P11);
\draw[ray,srcc] (Cj) -- (P10);
\draw[ray,srcc] (Cj) -- (P01);

\draw[planec!45!black,line width=0.35pt]
  (P00)--(P10)--(P11)--(P01)--cycle;
\foreach \a in {-\q,...,\q}{
  \foreach \b in {-\q,...,\q}{
    \node[smp,fill=black!80] at
      ($(X)+\a*\h*(E1)+\b*\h*(E2)$) {};
  }
}

\draw[vec,planec!40!black] (X) -- ($(X)+1.05*(E1)$)
  node[right=-1pt,yshift=-1pt] {$\bm e_1$};
\draw[vec,planec!40!black] (X) -- ($(X)+1.05*(E2)$)
  node[above right=-3pt and -1pt] {$\bm e_2$};
\draw[vec,black] (X) -- ($(X)+1.35*(N)$)
  node[above=-1pt] {$\bm n_\theta(\bm x)$};

\node[smp,fill=xc,minimum size=3.4pt] at (X) {};
\node[xc,inner sep=0pt] at
  ($(X)-0.5*\h*(E1)-0.45*\h*(E2)$) {$\bm x$};

\coordinate (H1) at ($(X)+0*\h*(E1)-3.1*\h*(E2)$);
\coordinate (H2) at ($(X)+1*\h*(E1)-3.1*\h*(E2)$);
\draw[{Bar[width=3pt]}-{Bar[width=3pt]},line width=0.4pt]
  (H1) -- (H2);
\node[below=0pt] at ($(H1)!0.5!(H2)$) {$h_i(\bm x)$};

\newcommand{\camera}[3]{%
  \coordinate (#2D) at ($(#1)!1.15cm!(X)$);
  \coordinate (#2U) at ($(#2D)!0.50cm!90:(#1)$);
  \coordinate (#2V) at ($(#2D)!0.50cm!-90:(#1)$);
  \coordinate (#21) at ($(#2U)+(0.05,0.26)$);
  \coordinate (#22) at ($(#2V)+(0.05,0.26)$);
  \coordinate (#23) at ($(#2V)-(0.05,0.26)$);
  \coordinate (#24) at ($(#2U)-(0.05,0.26)$);
  \foreach \k in {1,...,4} \draw[cam,#3] (#1) -- (#2\k);
  \filldraw[cam,#3,fill=#3!8]
    (#21)--(#22)--(#23)--(#24)--cycle;
  \fill[#3] (#1) circle (1.4pt);
}
\camera{Ci}{I}{refc}
\camera{Cj}{J}{srcc}

\filldraw[refc,line width=0.45pt,fill=refc!35]
  ($(ID)+(-0.10,-0.07)$) -- ($(ID)+(0.10,-0.05)$) --
  ($(ID)+(0.07,0.08)$) -- ($(ID)+(-0.11,0.06)$) -- cycle;
\filldraw[srcc,line width=0.45pt,fill=srcc!35]
  ($(JD)+(-0.13,-0.04)$) -- ($(JD)+(0.06,-0.11)$) --
  ($(JD)+(0.12,0.07)$) -- ($(JD)+(-0.05,0.12)$) -- cycle;

\node[refc,above left=0pt and -1pt] at (Ci) {$\bm c_i$};
\node[refc,left=1pt] at (I4) {$I_i$};
\node[srcc,above right=0pt and -1pt] at (Cj) {$\bm c_j$};
\node[srcc,right=1pt] at (J2) {$I_j$};

\def\pix{0.42}
\def\rhov{1}

\newcommand{\pixgrid}[1]{%
  \fill[#1!4]
    (-2.5*\pix,-2.5*\pix) rectangle (2.5*\pix,2.5*\pix);
  \foreach \t in {-1.5,-0.5,0.5,1.5}{
    \draw[gray!45,line width=0.3pt]
      (\t*\pix,-2.5*\pix) -- (\t*\pix,2.5*\pix);
    \draw[gray!45,line width=0.3pt]
      (-2.5*\pix,\t*\pix) -- (2.5*\pix,\t*\pix);
  }
  \draw[#1,line width=0.6pt]
    (-2.5*\pix,-2.5*\pix) rectangle (2.5*\pix,2.5*\pix);
}

\begin{scope}[shift={(-4.6,0.35)}]
  \pixgrid{refc}
  \def\AxI{0.90}\def\AyI{0.10}
  \def\BxI{-0.12}\def\ByI{0.72}
  \def\sI{\rhov*\pix}
  \filldraw[refc!80!black,line width=0.4pt,fill=refc!18]
    ({(-2*\AxI-2*\BxI)*\sI},{(-2*\AyI-2*\ByI)*\sI}) --
    ({( 2*\AxI-2*\BxI)*\sI},{( 2*\AyI-2*\ByI)*\sI}) --
    ({( 2*\AxI+2*\BxI)*\sI},{( 2*\AyI+2*\ByI)*\sI}) --
    ({(-2*\AxI+2*\BxI)*\sI},{(-2*\AyI+2*\ByI)*\sI}) -- cycle;
  \foreach \a in {-2,...,2}{
    \foreach \b in {-2,...,2}{
      \node[smp,fill=refc!75!black] at
        ({(\a*\AxI+\b*\BxI)*\sI},
         {(\a*\AyI+\b*\ByI)*\sI}) {};
    }
  }
  \node[smp,fill=xc,minimum size=3pt] at (0,0) {};
  \node[below=1pt,refc] at (0,-2.5*\pix) {$A_i(\bm x)$};
  \coordinate (insI) at (0,2.5*\pix);
\end{scope}
\draw[zoom,refc!70] (insI) -- ($(ID)+(-0.06,-0.09)$);

\begin{scope}[shift={(5.0,0.35)}]
  \pixgrid{srcc}
  \def\Ax{0.58}\def\Ay{-0.17}
  \def\Bx{0.22}\def\By{0.50}
  \def\s{\rhov*\pix}
  \filldraw[srcc!80!black,line width=0.4pt,fill=srcc!18]
    ({(-2*\Ax-2*\Bx)*\s},{(-2*\Ay-2*\By)*\s}) --
    ({( 2*\Ax-2*\Bx)*\s},{( 2*\Ay-2*\By)*\s}) --
    ({( 2*\Ax+2*\Bx)*\s},{( 2*\Ay+2*\By)*\s}) --
    ({(-2*\Ax+2*\Bx)*\s},{(-2*\Ay+2*\By)*\s}) -- cycle;
  \foreach \a in {-2,...,2}{
    \foreach \b in {-2,...,2}{
      \node[smp,fill=srcc!75!black] at
        ({(\a*\Ax+\b*\Bx)*\s},
         {(\a*\Ay+\b*\By)*\s}) {};
    }
  }
  \node[smp,fill=xc,minimum size=3pt] at (0,0) {};
  \node[below=1pt,srcc] at (0,-2.5*\pix) {$A_j(\bm x)$};
  \coordinate (insJ) at (0,2.5*\pix);
\end{scope}
\draw[zoom,srcc!70] (insJ) -- ($(JD)+(0.0,-0.10)$);

\end{tikzpicture}
}
\vspace{-6mm}
\caption{
At the first surface intersection $\bm x$, a $P\times P$ patch is sampled on the tangent plane with spacing $h_i(\bm x)$. Its samples are projected into the reference and visible source views, where the sampled colors are compared using normalized correlation.
}
\label{fig:tangent_patch}
\end{figure}
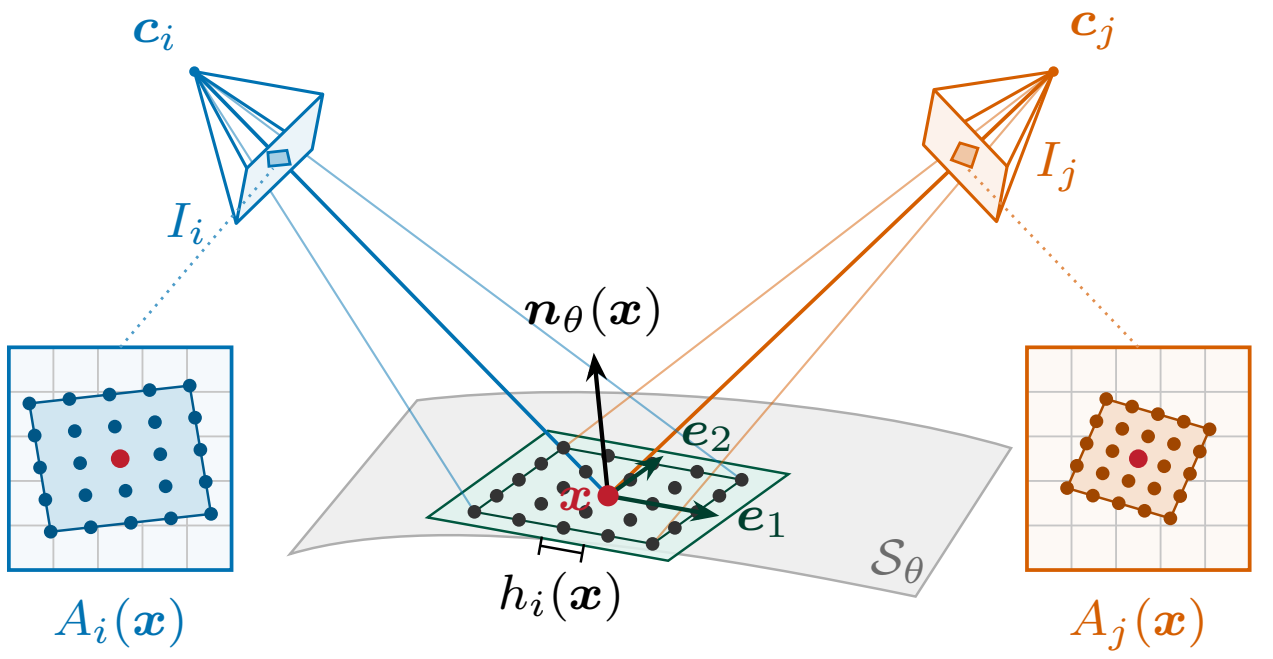

\smallskip\noindent\textbf{Patch photoconsistency.}
During optimization we sample a batch of reference image-pixel pairs
\(\mathcal{R} = \{(i,p)\}\) and use the valid subset
\(\mathcal{R}_\theta\subseteq\mathcal{R}\) of rays for which sphere tracing converges. To evaluate photometric agreement, we sample a planar \(P\times P\) patch in the tangent plane at \(x\) with $P=2q+1$. Let $(e_1(x),e_2(x))$ be an
orthonormal basis of $n_\theta(x)^\perp$ (Appendix~\ref{sec:orthonormal-basis}). We define the tangent-plane patch as
\begin{equation}
\mathcal{P}_x^i =
\big(x+h_i(x)(a\,e_1(x)+b\,e_2(x))\big)_
{(a,b)\in\llbracket-q,q\rrbracket^2}
\in \mathbb{R}^{3\times P\times P},
\end{equation}
where $h_i(x)=\rho z_i(x)/f_{x,i}$,
with $z_i(x)$ the camera-space depth and $f_{x,i}$
the focal length. Under the pinhole camera model, this yields a nominal spacing of \(\rho\) pixels for a fronto-parallel plane
(Fig.~\ref{fig:tangent_patch}). Sampling the patch on the surface itself, by tracing $P^2$ neighboring
rays, would capture its curvature but requires $P^2$ sphere-tracing runs per pixel.
On our reconstructed models, this alternative changes the normalized correlation by only $0.005$
on average, even where the tangent plane departs from the surface by more than
a pixel.

\noindent
We compare the bilinearly sampled projections of the tangent patch across visible source views,
following PMVS~\cite{furukawa2010pmvs} and NeuralWarp~\cite{darmon2022neuralwarp}. Let $\mathcal{N}_i$ denote the $K$ source views whose camera
centers are closest to $c_i$ and
$\mathcal{N}_i(x)=\mathcal{N}_i \cap \mathcal{V}_i(x)$. For each
source view \(j\in\mathcal{N}_i(x)\), we project the tangent patch
\(\mathcal{P}_x^i\) and bilinearly sample the reference and neighboring image intensities
$A_i(x)
=
I_i\!\left(\pi_i(\mathcal{P}_x^i)\right)$ and
$A_j(x)
=
I_j\!\left(\pi_j(\mathcal{P}_x^i)\right)$.
We quantify their discrepancy via the opposite of the normalized correlation~\cite{forsyth2012computer}
(also called NCC or ZNCC) of the two patches:
\begin{equation}
\label{eq:zncc}
r_{ij}(x) = -\operatorname{corr}(A_i(x),A_j(x)),
\end{equation}
where $\operatorname{corr}(a,b) = (a-\bar a)^\top(b-\bar b)\,/\,\bigl(\|a-\bar a\|\,\|b-\bar b\|\bigr)$
is the normalized correlation.
Our photometric objective assumes approximate consistency of local intensity patterns across views. To mitigate the effects of specularities and other photometric inconsistencies, we follow PMVS~\cite{furukawa2010pmvs} and reject pairs with low photometric correlation, retaining
\begin{equation}
\mathcal{N}^*_i(x)
=
\left\{
j\in\mathcal{N}_i(x)
\;\middle|\;
r_{ij}(x)<\tau
\right\}.
\end{equation}
To lighten notation, we omit the dependence of 
 $\mathcal{N}^*_i(x)$ and $r_{ij}(x)$ on $\theta$. The objective we optimize is 
\begin{equation}
\label{eq:photoloss}
\mathcal{L}_{\text{photo}}(\theta)
=
\mathbb{E}_{(i,p)\in\mathcal{R}^*_\theta}
\left[
\frac{1}{|\mathcal{N}^*_i(x)|}
\sum_{j\in\mathcal{N}^*_i(x)}
r_{ij}(x)
\right],
\end{equation}
where
$\mathcal{R}^*_\theta
=
\left\{
(i,p)\in\mathcal{R}_\theta
:
|\mathcal{N}^*_i(x_\theta(i,p))|>0
\right\}.$

\subsection{Sphere tracing}
Evaluating the photometric objective requires computing first surface intersections along camera rays. 
For this, we use the
sphere tracing algorithm described below~\cite{hart1996sphere}. For a \(1\)-Lipschitz MLP, the algorithm converges
linearly to \(t_\theta(i,p)\) at a non-grazing intersection. Converged intersections are refined with two safeguarded Newton steps along the ray.

\begin{algorithm}[H]
\caption{Sphere tracing for $f_\theta$}
\begin{algorithmic}[1]
\Require $i,p,\varepsilon,T,M$
\State $t \gets 0$
\For{$k=1,\ldots,M$}
    \State $x \gets c_i+t\,u_i(p)$; $s \gets f_\theta(x)$
    \State \textbf{if} $|s|\leq\varepsilon$ \textbf{then return} $x$
    \State \textbf{if} $t\geq T$ \textbf{then return} $\varnothing$
    \State $t \gets \min(t+|s|,T)$
\EndFor
\State \textbf{return} $\varnothing$
\end{algorithmic}
\end{algorithm}

\par\smallskip\noindent\textbf{Implicit differentiation.}
Following prior work~\cite{yariv2020idr,niemeyer2020dvr,liu2020dist},
we differentiate the first-hit point
$x=x_\theta(i,p)$ with respect to $\theta$
via the implicit function theorem.
At an intersection satisfying
$f_\theta(x)=0$ and
$\nabla_x f_\theta(x)^\top u_i(p)\neq 0$, we have
\begin{equation}
\label{eq:intersection-gradient}
\nabla_\theta x_\theta(i,p)
=
-\frac{1}{\nabla_x f_\theta(x)^\top u_i(p)}
\,\nabla_\theta f_\theta(x)\,u_i(p)^\top.
\end{equation}
This avoids backpropagating through sphere tracing iterations.

\par\smallskip\noindent\textbf{Optimization.}
At each iteration, accepted source views and patch orientations are recomputed and held fixed during backpropagation. View selection is locally constant almost everywhere, while detaching patch orientations avoids second-order derivatives (Appendix~\ref{sec:computational}). For $\ell_{ij}(\theta)=r_{ij}(x_\theta(i,p))$, differentiation with fixed patch orientations and Eq.~\eqref{eq:intersection-gradient} yield

\begin{equation}
\nabla_\theta\ell_{ij}(\theta)
=
-\frac{
\nabla_x r_{ij}(x)^\top u_i(p)
}{
\nabla_x f_\theta(x)^\top u_i(p)
}
\nabla_\theta f_\theta(x).
\end{equation}

\subsection{1-Lipschitz architecture}
To ensure convergence of sphere tracing, we constrain $f_\theta$ to be $1$-Lipschitz. We adopt a backbone $\Phi_\theta:\mathbb{R}^D\to\mathbb{R}^D$
interleaving Convex Potential Layers (CPL) and MaxMin
activations~\citep{anil2019groupsort,meunier2022cpl,prach2024layers}, where $D$ is the even dimension of the hidden layers. The resulting MLP is parameterized as
\begin{equation}
f_\theta(x)=v^\top\Phi_\theta(Ax)+\beta,
\end{equation}
where $A\in\mathbb{R}^{D\times 3}$ satisfies
$A^\top A=I_3$,
$v=\widetilde{v}/\|\widetilde{v}\|_2$ and
$\beta\in\mathbb{R}$.
Since each component is $1$-Lipschitz, so is $f_\theta$.
Further architectural details are provided in
Appendix~\ref{sec:architecture-details}.

\subsection{Positional encoding}
To capture finer geometric details, we lift the input with Fourier positional encoding, widely used in neural surface reconstruction~\citep{mildenhall2020nerf,
tancik2020fourier}.
For $L\geq1$,
\begin{equation}
\gamma_L(x)
=
\left[
x,
\left(
\sin(2^k x_j),\cos(2^k x_j)
\right)_{\substack{j=1,\ldots,3\\k=0,\ldots,L-1}}
\right].
\label{eq:positional-encoding}
\end{equation}
The implicit field becomes
$f_\theta :x\mapsto v^\top\Phi_\theta(A\gamma_L(x))+\beta$ with \(A\in\mathbb{R}^{D\times(6L+3)}\) and $\Lip(f_\theta)\leq \Lip(\gamma_L)
=
(1+(4^L-1) /3)^{1/2}$ for the \(\ell_2\) norm.
This variant is therefore no longer formally 1-Lipschitz.
Dividing sphere tracing steps by $\operatorname{Lip}(\gamma_L)$
yields highly conservative steps. Following~\cite{lenain2026surface},
we instead rescale $f_\theta$ every 50 iterations by dividing
it by an empirical estimate of its Lipschitz constant,
computed as the maximum gradient norm over 32k points
sampled in the tracing domain. We assess sphere tracing reliability against interval-certified
first-hit references on 100k DTU rays per model.
With $L=6$ and rescaling, first-hit accuracy reaches
99.83\%, with no observed
overshoots (Table~\ref{tab:tracing-ablation}).

\begin{table*}[t]
\centering
\caption{
Surface reconstruction results on BMVS ($\mathrm{CD}_1 \times 10^{3}$).
Each scene reports $\mathrm{CD}_1$ with the standard deviation of its point-to-surface
distances in parentheses. The last column reports the mean $\pm$ standard deviation
over the 8 scenes. $^\dagger$ indicates the use of foreground masks.
The top three results among neural implicit and Gaussian splatting methods
are highlighted in red, orange, and yellow, respectively.
}
\vspace{-2mm}
\label{tab:blendedmvs-reference}

\begingroup
\fontsize{8}{9.4}\selectfont
\setlength{\tabcolsep}{2pt}
\renewcommand{\arraystretch}{0.90}

\begin{tabular*}{\textwidth}{
@{}l@{\extracolsep{\fill}}*{8}{r}!{\hspace{6pt}}r@{}
}
\toprule
\textbf{Method}
& \textbf{Dog}
& \textbf{Bear}
& \textbf{Clock}
& \textbf{Durian}
& \textbf{Man}
& \textbf{Sculpture}
& \textbf{Stone}
& \textbf{Jade}
& \textbf{Mean} \\
\midrule

\multicolumn{10}{l}{\textit{Classical MVS}} \\
PMVS$_{\mathrm{trim}=7}$
& 1.91\pp{3.97} & 2.08\pp{3.30} & 3.41\pp{4.57} & 2.99\pp{4.34} & 1.69\pp{3.40} & 1.85\pp{3.48} & 1.98\pp{2.99} & 5.79\pp{6.02} & 2.71\sd{1.38} \\

COLMAP$_{\mathrm{trim}=7}$
& 2.11\pp{4.39} & 1.51\pp{2.58} & 2.70\pp{4.05} & 3.21\pp{4.43} & 1.38\pp{2.68} & 2.45\pp{4.29} & 2.09\pp{3.14} & 6.23\pp{6.48} & 2.71\sd{1.54} \\

ACMMP$_{\mathrm{trim}=7}$
& 1.80\pp{4.14} & 1.37\pp{2.65} & 2.55\pp{4.54} & 3.08\pp{4.12} & 1.51\pp{3.04} & 1.86\pp{3.39} & 1.59\pp{3.16} & 6.58\pp{6.40} & 2.54\sd{1.73} \\

\midrule
\multicolumn{10}{l}{\textit{Neural implicit surfaces}} \\
IDR$^\dagger$
& 3.48\pp{5.04} & 4.97\pp{6.26} & 2.64\pp{3.89} & 10.36\pp{7.11} & 2.86\pp{4.19} & 2.40\pp{3.83} & 4.56\pp{5.91} & 6.84\pp{6.49} & 4.76\sd{2.70} \\

NeuS$^\dagger$
& 2.83\pp{4.65} & 4.38\pp{5.24} & 2.36\pp{3.39} & 5.37\pp{6.25} & 2.27\pp{3.74} & 2.76\pp{4.00} & 2.81\pp{4.07} & 5.42\pp{5.98} & 3.52\sd{1.32} \\

NeuralWarp
& 1.95\pp{3.99} & 1.53\pp{2.60} & 2.74\pp{2.99} & 3.49\pp{4.69} & 3.30\pp{5.20} & 4.72\pp{5.32} & \third{1.85}\pp{3.12} & 7.15\pp{6.51} & 3.34\sd{1.86} \\

ProbeSDF$^\dagger$
& 2.45\pp{4.83} & 2.04\pp{3.89} & \third{1.68}\pp{3.56} & 3.73\pp{5.52} & 1.94\pp{3.75} & \second{1.60}\pp{3.57} & \best{1.39}\pp{3.18} & \second{4.01}\pp{5.71} & \third{2.36}\sd{0.99} \\

Geo-NeuS
& \second{1.60}\pp{3.45} & \best{1.19}\pp{2.57} & \best{1.26}\pp{2.32} & 2.90\pp{4.54} & \second{1.68}\pp{3.36} & 2.00\pp{3.60} & 1.86\pp{3.26} & \third{4.88}\pp{5.74} & \second{2.17}\sd{1.22} \\

\textbf{Ours}
& \third{1.90}\pp{4.34} & \third{1.39}\pp{2.84} & \second{1.46}\pp{2.22} & \third{2.75}\pp{4.23} & \best{1.57}\pp{3.64} & \best{1.51}\pp{2.98} & \second{1.72}\pp{3.03} & \best{3.89}\pp{5.05} & \best{2.02}\sd{0.87} \\

\midrule
\multicolumn{10}{l}{\textit{Gaussian splatting}} \\
2DGS
& 4.61\pp{5.97} & 12.32\pp{7.26} & 11.03\pp{7.24} & 6.60\pp{7.12} & 4.37\pp{5.53} & 12.39\pp{7.31} & 4.32\pp{6.12} & 4.92\pp{5.54} & 7.57\sd{3.69} \\

PGSR
& 2.29\pp{4.24} & 1.51\pp{2.22} & 2.92\pp{4.22} & \best{2.66}\pp{3.75} & 1.96\pp{3.38} & 2.04\pp{2.89} & 1.98\pp{2.59} & 5.10\pp{5.25} & 2.56\sd{1.12} \\

AmbiSuR
& \best{1.57}\pp{3.08} & \second{1.29}\pp{2.04} & 3.21\pp{4.95} & \second{2.68}\pp{3.96} & \second{1.68}\pp{2.85} & \third{1.89}\pp{2.54} & 1.93\pp{3.04} & 5.27\pp{5.45} & 2.44\sd{1.30} \\

\bottomrule
\end{tabular*}
\endgroup
\end{table*}

\begin{figure*}[t]
\vspace{-2mm}
    \centering
    \includegraphics[width=\textwidth]{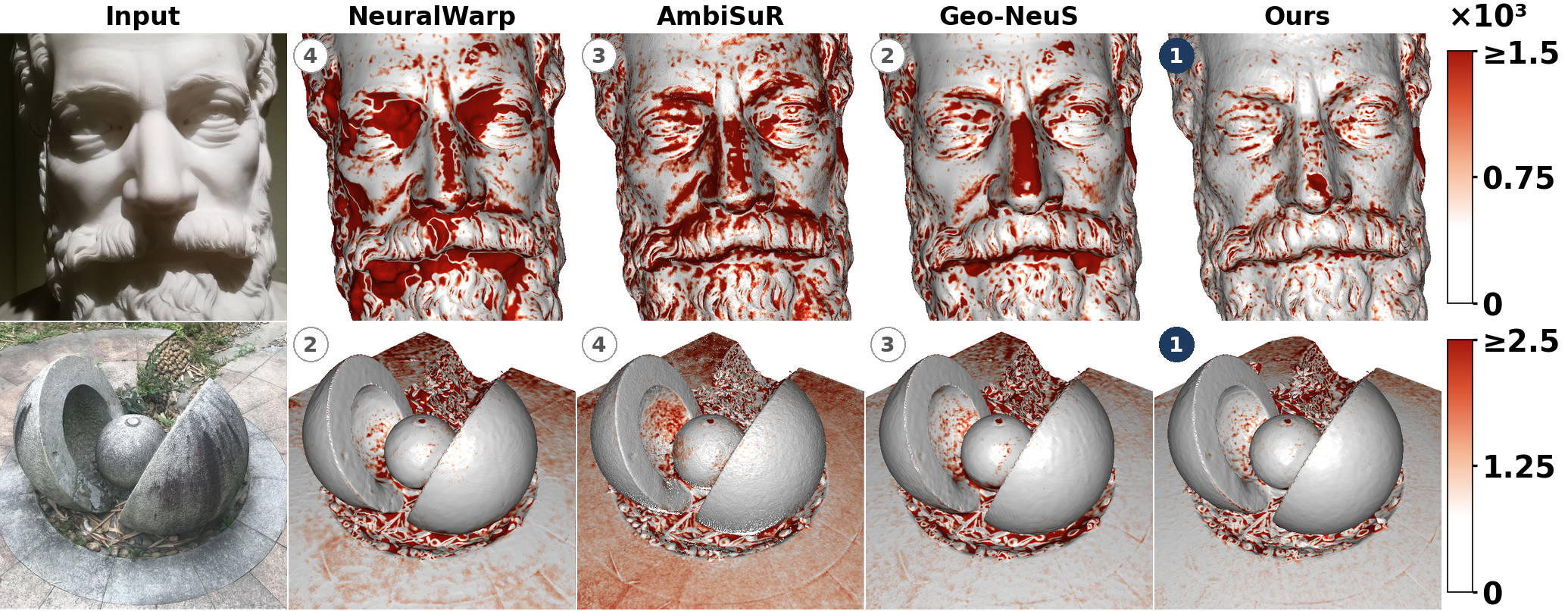}
    \vspace{-6mm}
    \caption{Surface reconstruction results on two BMVS scenes. Colors indicate distances to the reference mesh, scaled by $10^3$. Each scene uses the same color scale across methods, with red indicating larger errors. Circled
numbers rank methods by Chamfer distance.}
    \label{fig:bmvs_error}
\end{figure*}

\subsection{Initialization}

Patch photoconsistency requires sufficiently accurate initial correspondences. Like NeuralWarp~\citep{darmon2022neuralwarp}, which introduces patch warping only after an initial VolSDF~\citep{yariv2021volsdf} fitting phase, we initialize $f_\theta$ from a coarse geometric proxy.
Starting from an enclosing sphere, we carve away free space using depth
maps from the classical MVS method ACMMP~\cite{xu2022acmmp}, then fit $f_\theta$ to the SDF of the resulting volume (Appendix~\ref{app:init}).

\section{Experiments}

\subsection{Implementation details}

Unless stated otherwise, we use an MLP with 8 layers of width 256, Convex Potential Layers, MaxMin activations and positional encoding with $L=6$ (Appendix~\ref{sec:architecture-details}). We optimize for 300k steps using Adam with cosine learning rate decay from $10^{-3}$ to $10^{-4}$, sampling 16,384 rays per batch, with 90\% drawn uniformly from pixels
whose rays intersect the initial surface and 10\% from
the remaining pixels. For each reference view, we select $K=6$ source views and compute normalized correlation on grayscale intensities sampled over $5\times5$ tangent patches, with $\rho=1$. Further implementation details appear in Appendix~\ref{app:st-params}.

\paragraph{Baselines.}
We compare the proposed method with neural implicit surface methods IDR \citep{yariv2020idr},
NeuS \citep{wang2021neus}, NeuralWarp \citep{darmon2022neuralwarp}, ProbeSDF \citep{toussaint2025probesdf} and Geo-NeuS \citep{fu2022geoneus},
classical MVS methods COLMAP \citep{schoenberger2016pixelwise}, ACMMP \citep{xu2022acmmp} and
PMVS \citep{furukawa2010pmvs}, and Gaussian-splatting surface reconstruction
methods 2DGS \citep{huang2024_2dgs}, PGSR \citep{chen2025pgsr},
and AmbiSuR \citep{li2026ambisur}.
IDR, NeuS and ProbeSDF use
foreground masks during fitting, while AmbiSuR uses pretrained depth priors. For classical MVS baselines, we apply Poisson surface reconstruction~\citep{kazhdan2006poisson} with density trimming at threshold 7, following~\citep{fu2022geoneus}.
We report Neuralangelo~\citep{li2023neuralangelo} separately in Appendix~\ref{app:neuralangelo_bmvs}, as our runs with its released implementation yielded substantially worse results than those reported in the original paper (1.71 vs. 0.61 mm).

\subsection{BMVS}
To complement DTU's fixed camera trajectory (Sec.~\ref{sec:dtu}), we evaluate
on BMVS~\cite{yao2020blendedmvs}, whose object-centric scenes provide
approximately $360^\circ$ view coverage, testing our method
under more varied visibility configurations.
 We evaluate on \(8\) scenes used in
prior neural surface reconstruction benchmarks, with \(31\) to \(143\)
calibrated images at \(768\times576\) resolution. Unlike DTU, BMVS uses reference meshes reconstructed from images,
without independently measured geometry or reported accuracy
~\cite{yao2020blendedmvs}. We follow the BMVS
protocol used by ProbeSDF~\cite{toussaint2025probesdf}. Neural implicit meshes are extracted using marching cubes at resolution $512^3$~\cite{lorensen1987marching}. The metric is the
bidirectional unsigned point-to-surface distance between the predicted mesh and
the reference mesh. Distances larger than \(0.025\) in normalized
scene coordinates are discarded before averaging, and scores are reported \( \times 10^3\).
\\
As shown in Table~\ref{tab:blendedmvs-reference}, our method achieves the lowest mean error among all compared methods (2.02 versus 2.17 for Geo-NeuS), outperforming Geo-NeuS on five of eight scenes. The largest improvement occurs on Jade ($3.89$ vs $4.88$), which combines the highest occlusion rate (Fig.~\ref{fig:occlusion_prevalence}) with intricate geometry. Removing occlusion filtering increases error on Jade to $4.20$, while enlarging patches from $5\times5$ to the $11\times11$ size used in Geo-NeuS raises it to $5.58$. This is consistent with larger patches being less well approximated by tangent planes on curved surfaces and more likely to cross depth discontinuities near thin structures.

\subsection{DTU}
\label{sec:dtu}

DTU~\cite{jensen2014large,anaes2016large} provides calibrated multi-view images with structured light ground truth, for which Jensen et al.~\cite{jensen2014large} report a standard deviation of
0.14\,mm, measured on a sphere of known geometry. We evaluate the standard 15 scans, each containing 49 to 64 images at $1600\times1200$ resolution. We reran baselines using the official implementations and recovered their published results under their original evaluation protocols, except for Geo-NeuS (0.60 mm versus the published 0.51 mm), consistent with independent reruns reporting 0.59--0.62 mm~\citep{jiang2023coordinate,ren2024improving}. For both rerun columns, all meshes undergo the same silhouette crop using masks dilated by 12 pixels~\cite{darmon2022neuralwarp}. Neural implicit meshes are extracted using marching cubes at resolution $512^3$~\cite{yariv2020idr,wang2021neus}, while Gaussian splatting methods retain their mesh extraction procedures. We sample the meshes, transform the samples to DTU coordinates, and compute bidirectional Chamfer distance $\mathrm{CD}_1$ (mm) against the official ground truth (Appendix~\ref{sec:eval-metrics}). 

\begin{figure*}[!ht]
    \centering
    \includegraphics[width=\textwidth]{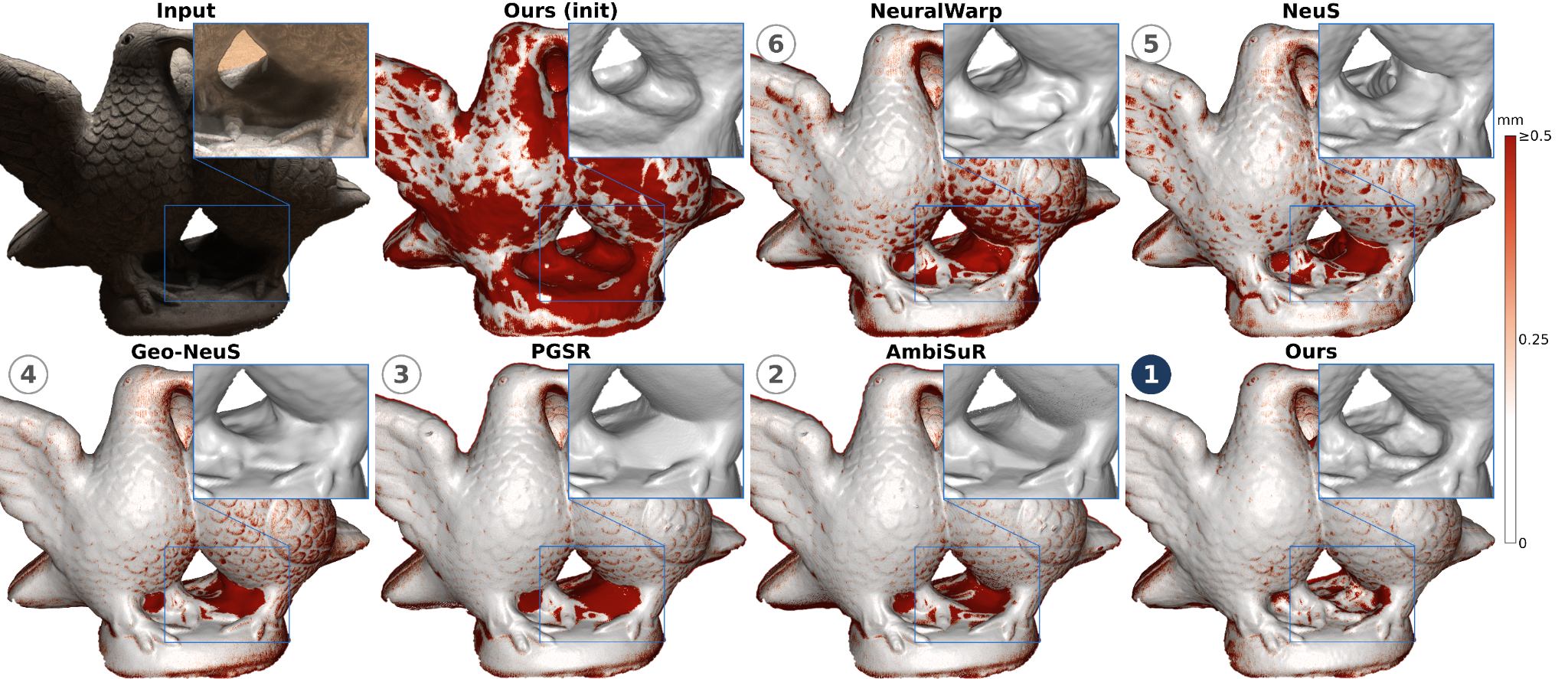}
    \vspace{-6mm}
    \caption{Surface reconstruction on DTU scan 106. Insets highlight the bird's leg, where our method preserves fine geometric details. Red indicates distance to the ground truth surface, saturated at 0.5 mm. Circled
numbers rank methods by $\mathrm{CD}_1$ after visibility culling.
 }
    \label{fig:dtu_qualitative}
\end{figure*}

\begin{table}[!ht]
\centering
\caption{
Surface reconstruction results on DTU (mean $\mathrm{CD}_1$ in mm
$\pm$ standard deviation over the 15 scans).
We report published results alongside reruns of the authors’ released implementations. All reruns use the same 12-pixel silhouette crop, with results reported before and after visibility culling.
$^\dagger$ indicates the use of foreground masks.
Per-scan results are in Appendix~\ref{sec:additional-results}.
The top three results in each column are
highlighted in red, orange, and yellow, respectively.
}
\label{tab:dtu-results}
\vspace{-2mm}
\small
\setlength{\tabcolsep}{6pt}
\renewcommand{\arraystretch}{1}

\begin{tabular}{lccc}
\toprule
\textbf{Method}
& \textbf{Original}
& \textbf{Rerun}
& \shortstack{\textbf{Rerun + culling}} \\
\midrule

\multicolumn{4}{l}{\textit{Classical MVS}} \\
PMVS$_{\mathrm{trim}=7}$   & 0.84\sd{0.41} & 0.84\sd{0.41} & 0.78\sd{0.38} \\
COLMAP$_{\mathrm{trim}=7}$ & 0.66\sd{0.31} & 0.66\sd{0.31} & 0.57\sd{0.26} \\
ACMMP$_{\mathrm{trim}=7}$  & 0.61\sd{0.29} & 0.60\sd{0.29} & 0.54\sd{0.26} \\

\midrule
\multicolumn{4}{l}{\textit{Neural implicit surfaces}} \\
IDR$^\dagger$   & 0.90\sd{0.43} & 0.91\sd{0.44} & 0.79\sd{0.41} \\
NeuS$^\dagger$  & 0.77\sd{0.36} & 0.79\sd{0.34} & 0.68\sd{0.29} \\
NeuralWarp      & 0.68\sd{0.24} & 0.68\sd{0.24} & 0.61\sd{0.21} \\
ProbeSDF$^\dagger$
                & 0.68\sd{0.29} & 0.70\sd{0.32} & 0.60\sd{0.28} \\
Geo-NeuS        & \second{0.51}\sd{0.22} & 0.60\sd{0.26} & \third{0.50}\sd{0.21} \\
\textbf{Ours}   & 0.57\sd{0.25} & \third{0.57}\sd{0.25} & \second{0.49}\sd{0.21} \\

\midrule
\multicolumn{4}{l}{\textit{Gaussian splatting}} \\
2DGS            & 0.80\sd{0.34} & 0.81\sd{0.44} & 0.75\sd{0.41} \\
PGSR            & \third{0.52}\sd{0.21} & \second{0.55}\sd{0.23} & 0.51\sd{0.22} \\
AmbiSuR         & \best{0.46}\sd{0.18} & \best{0.49}\sd{0.22} & \best{0.48}\sd{0.22} \\

\bottomrule
\end{tabular}
\end{table}

\paragraph{Visibility culling.}
Neural implicit methods mesh the zero level set of the full field, whereas
Gaussian methods fuse depth maps rendered from the training
views~\citep{huang2024_2dgs, chen2025pgsr}. Consequently, neural implicit meshes can retain more spurious geometry
in poorly observed regions (Fig.~\ref{fig:culling_visualization}).
We therefore apply the same visibility culling to all rerun meshes. Using only the meshes
and input cameras, a face is kept if it is unoccluded in at least five cameras
whose viewing rays make at most $75^\circ$ with its normal
(Appendix~\ref{sec:visibility-cull}), similar to~\citep{weidenfeller2026adele}. Culling mainly removes poorly observed peripheral regions (Figure~\ref{fig:culling_visualization}).
\noindent
For our method, culling improves the accuracy term from 0.60 to
0.43 mm while leaving completeness nearly unchanged
(0.54 to 0.56 mm), suggesting the removed regions
contribute geometric errors without substantially improving
surface coverage. The resulting mean $\mathrm{CD}_1$ of 0.49 mm
nearly matches AmbiSuR (0.48 mm), which additionally
uses pretrained depth priors. Fig.~\ref{fig:visibility_culling} reports Chamfer sensitivity to the culling parameters.

\section{Discussion}

\subsection{Ablations} 
\begin{table}[!ht]
\centering
\caption{Ablations on DTU (mean $\mathrm{CD}_1$ in mm $\pm$ standard deviation
over the 15 scans). Each variant modifies a component of our default
configuration (Ours). The RGB pixels variant replaces grayscale intensities
with RGB values.}
\label{tab:ablations}
\small
\begin{tabular}{lc}
\toprule
\textbf{Variant} & \textbf{CD$_1$ $\downarrow$} \\
\midrule
No positional encoding & 0.98\sd{0.34} \\
Single-pixel $\ell_1$ & 0.85\sd{0.43} \\
RGB pixels & 0.62\sd{0.26} \\
No occlusion filtering & 0.61\sd{0.26} \\
NeuS MLP + eikonal & 0.60\sd{0.27} \\
\midrule
\textbf{Ours} & \textbf{0.57}\sd{0.25} \\
\bottomrule
\end{tabular}
\end{table}

\par\smallskip\noindent\textbf{Loss.} Replacing normalized correlation with single-pixel
$\ell_1$ increases mean CD$_1$ from 0.57 to 0.85\,mm (Table~\ref{tab:ablations}),
consistent with the depth localization results in
Fig.~\ref{fig:photometric_localization}.
In contrast, using the $\ell_1$ distance between normalized patches yields similar depth localization accuracy.

\begin{figure}[!ht]
    \centering
    \includegraphics[width=\linewidth]{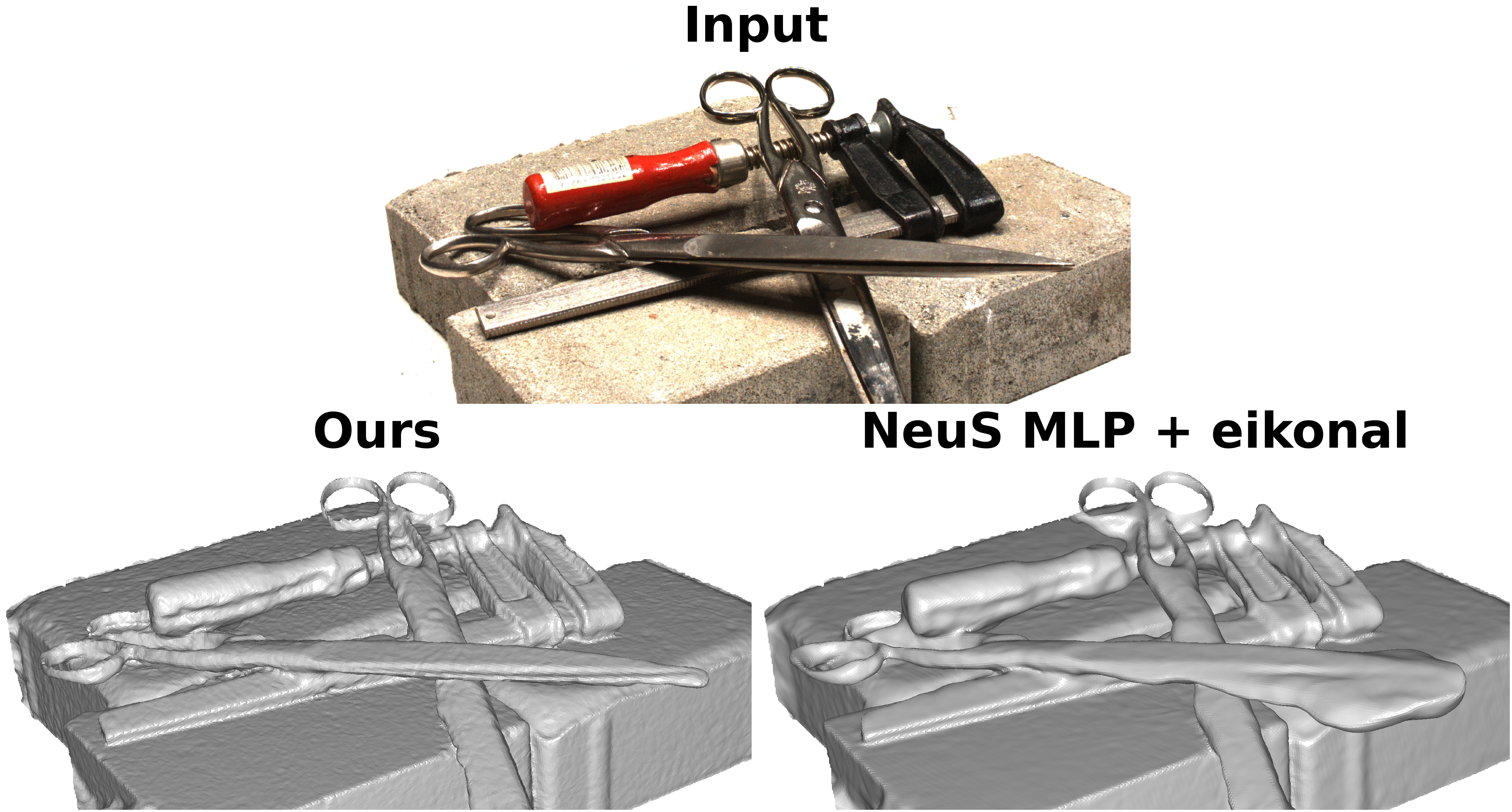}
    \vspace{-6mm}
    \caption{Surface reconstructions of DTU scan 37 using CPL and the NeuS MLP, both optimized with $\mathcal{L}_{\mathrm{photo}}$, with additional eikonal regularization for the latter.}
    \label{fig:architecture_ablation}
\end{figure}

\par\smallskip\noindent\textbf{Architecture.}
The unconstrained MLP architecture of NeuS, with eikonal
regularization, achieves a similar mean error
(0.60 vs. 0.57\,mm; Table~\ref{tab:ablations}) but produces smoother surfaces
at the expense of fine geometric details, as illustrated in Figure~\ref{fig:architecture_ablation} on DTU scan 37.
Our default method is somewhat rougher but better
preserves the fine structures of the scissors.

\par\smallskip\noindent\textbf{Positional encoding.}
Without positional encoding, the network captures the
coarse geometry but fails to recover fine surface details,
as illustrated in Figure~\ref{fig:pe_ablation} on DTU scan 122.
This loss of detail is reflected in the mean CD$_1$,
which increases from 0.57 to 0.98\,mm (Table~\ref{tab:ablations}).

\par\smallskip\noindent\textbf{Initialization.}
Initialization provides a coarse surface from which
meaningful photometric correspondences can be established.
Starting from the ACMMP carved proxy (1.55\,mm), our method
achieves 0.57\,mm, versus 1.28\,mm for classical photometric
mesh refinement~\cite{vu2012highaccuracy}.
Starting instead from a more accurate Poisson ACMMP mesh
(0.61\,mm), classical refinement yields 0.62\,mm
(Table~\ref{tab:classical-refinement}).
For our method, replacing ACMMP with COLMAP yields a similar
error (0.61\,mm), suggesting that the final accuracy depends
more on the subsequent surface optimization than on the
particular MVS initializer.
Starting from a sphere still achieves 0.85\,mm,
with the largest degradations on thin structures
and cavities. Detailed results are given in
Table~\ref{tab:init_ablation}.

\subsection{Visibility and depth localization}
To isolate the effect of visibility filtering, we fix the
reconstructed surface and estimate depth along reference rays
by minimizing the patch photoconsistency loss, with
and without source-view occlusion filtering.
We compare these estimates against depths derived from
DTU ground-truth point clouds.
Partially occluded rays are those whose ground-truth surface
points are visible from at least two and occluded from at least
two of the six selected source views.
Across 15 DTU scans, filtering reduces mean absolute depth
error from 2.81 to 2.13\,mm on 83k partially occluded rays,
compared with 1.19 to 1.15\,mm on 148k reliable foreground
rays. Improvements across all 15 scenes indicate that
visibility filtering primarily benefits depth localization
under partial occlusion (Table~\ref{tab:visibility}).

\subsection{Patch photoconsistency vs. volume rendering}
To assess the effect of replacing volumetric rendering with
direct patch photoconsistency, we compare our formulation
with NeuS, which uses volumetric rendering without explicit
multi-view matching. We initialize NeuS from the same coarse
surface, retaining its architecture, rendering loss and
eikonal regularization, but without foreground masks.
With matched initialization and geometry architecture,
our formulation outperforms NeuS on 14 of 15 DTU scans
(0.60 vs.\ 0.75 mm), while achieving the same mean accuracy
as Geo-NeuS (0.60 mm), which combines volumetric rendering
with patch photoconsistency.

\begin{table}[!ht]
\centering
\small
\caption{Volumetric rendering versus patch photoconsistency on DTU (mean $\mathrm{CD}_1$ in mm). No visibility culling is applied.}
\label{tab:direct_vs_volumetric}
\vspace{-2mm}
\begin{tabular}{lllc}
\toprule
Formulation & Architecture & Initialization & CD$_1$ \\
\midrule
NeuS & NeuS MLP + eikonal & Sphere & 0.84 \\
NeuS & NeuS MLP + eikonal & ACMMP & 0.75 \\
Geo-NeuS & Geo-NeuS MLP & Sphere & 0.60 \\
Geo-NeuS & Geo-NeuS MLP & ACMMP  & 0.60 \\
Ours & NeuS MLP + eikonal & ACMMP & 0.60 \\
Ours & CPL MLP & ACMMP & \textbf{0.57} \\
\bottomrule
\end{tabular}
\end{table}

\subsection{Computational cost}
Sphere tracing makes the forward pass costly due to
iterative network evaluations and visibility queries,
whereas implicit differentiation keeps the backward
pass inexpensive. This yields a runtime of 5\,h\,10\,min per DTU scene
on an H100 for our optimized
implementation (Appendix~\ref{sec:efficient-st}), including 10\,min for initialization,
comparable to NeuS (4\,h) and Geo-NeuS (5\,h) and faster than Neuralangelo (7\,h), but slower than Gaussian splatting methods (10–20 min) (Table~\ref{tab:runtime}).

\section{Conclusion}
We have demonstrated that accurate neural surface reconstruction does not require joint geometry and appearance optimization. By optimizing first-hit surface intersections through multi-view patch photoconsistency, our formulation avoids the coupling between geometry and appearance inherent in rendering approaches while achieving competitive or superior geometric accuracy. These results suggest that direct surface optimization is an effective alternative to volumetric rendering with appearance models for neural surface reconstruction.
As an extension, photometric bundle adjustment~\cite{delaunoy2014photometric},
alternating between camera pose and surface optimization,
reduces mean BMVS CD$_1$ from 2.022 to 1.896 (6.2\%),
improving all eight scenes (Appendix~\ref{sec:bundle-adjust}).
Similar pose refinement would probably also benefit other reconstruction methods. This formulation also has limitations. Photoconsistency constrains only surface regions with valid intersections and multi-view correspondences, so the method depends on initialization, although similar results from ACMMP and COLMAP initializations show that it does not rely on a particular MVS method. Omitting geometric regularization isolates what photoconsistency alone can constrain, at the cost of local roughness in some reconstructions. Normalized correlation also discards absolute intensity and contrast, motivating richer photometric matching objectives. Finally, extending the approach to larger, unbounded scenes remains future work.

\section*{Acknowledgments}
This work was supported in part by the French government under management of Agence Nationale de la Recherche as part of the “France 2030” program, PR[AI]RIE-PSAI project, reference ANR-23- IACL-0008. J.P. was supported in part by the Louis Vuitton/ENS chair in artificial intelligence, the Institute of Information \& Communications Technology Planning \& Evaluation (IITP) grant funded by the Korean Government (MSIT) (No. RS-2024-00457882, National AI Research Lab Project), and a Global Distinguished Professorship at the Courant Institute of Mathematical Sciences and the Center for Data Science at New York University.
{
\small
\nocite{zhang2020nerfpp,wang2026surfaces}
\bibliographystyle{ieeenat_fullname}
\bibliography{references}

@article{anaes2016large,
  author = {H. Aan{\ae}s and R. Jensen and G. Vogiatzis and E. Tola and A. Dahl},
  title = {{Large-scale data for multiple-view stereopsis}},
  journal = {International Journal of Computer Vision},
  volume = {120},
  number = {2},
  pages = {153--168},
  year = {2016},
}

@inproceedings{anil2019groupsort,
  author = {C. Anil and J. Lucas and R. Grosse},
  title = {{Sorting out Lipschitz function approximation}},
  booktitle = {ICML},
  year = {2019},
  pages = {291--301},
}

@inproceedings{bangaru2022differentiable,
  author = {S. P. Bangaru and M. Gharbi and T.-M. Li and F. Luan and K. Sunkavalli and M. Hasan and S. Bi and Z. Xu and G. Bernstein and F. Durand},
  title = {{Differentiable rendering of neural SDFs through reparameterization}},
  booktitle = {ACM SIGGRAPH Asia Conference Proceedings},
  year = {2022},
  pages = {1--9},
  note = {Article 22},
}

@article{chen2025pgsr,
  author = {D. Chen and H. Li and W. Ye and Y. Wang and W. Xie and S. Zhai and N. Wang and H. Liu and H. Bao and G. Zhang},
  title = {{PGSR: Planar-based Gaussian splatting for efficient and high-fidelity surface reconstruction}},
  journal = {IEEE Transactions on Visualization and Computer Graphics},
  volume = {31},
  number = {9},
  pages = {6100--6111},
  year = {2025},
}

@article{coiffier2024lipschitz,
  author = {G. Coiffier and L. B{\'e}thune},
  title = {{1-Lipschitz neural distance fields}},
  journal = {Computer Graphics Forum},
  volume = {43},
  number = {5},
  pages = {e15128},
  year = {2024},
}

@inproceedings{darmon2022neuralwarp,
  author = {F. Darmon and B. Bascle and J.-C. Devaux and P. Monasse and M. Aubry},
  title = {{Improving neural implicit surfaces geometry with patch warping}},
  booktitle = {CVPR},
  year = {2022},
}

@inproceedings{delaunoy2014photometric,
  author = {A. Delaunoy and M. Pollefeys},
  title = {{Photometric bundle adjustment for dense multi-view 3D modeling}},
  booktitle = {CVPR},
  year = {2014},
  pages = {1486--1493},
}

@inproceedings{faugeras1998complete,
  author = {O. Faugeras and R. Keriven},
  title = {{Complete dense stereovision using level set methods}},
  booktitle = {ECCV},
  year = {1998},
  pages = {379--393},
}

@book{forsyth2012computer,
  author = {D. A. Forsyth and J. Ponce},
  title = {{Computer Vision: A Modern Approach}},
  publisher = {Pearson},
  edition = {2nd},
  year = {2012},
}

@inproceedings{fu2022geoneus,
  author = {Q. Fu and Q. Xu and Y.-S. Ong and W. Tao},
  title = {{Geo-NeuS: Geometry-consistent neural implicit surfaces learning for multi-view reconstruction}},
  booktitle = {NeurIPS},
  year = {2022},
}

@article{furukawa2010pmvs,
  author = {Y. Furukawa and J. Ponce},
  title = {{Accurate, dense, and robust multiview stereopsis}},
  journal = {IEEE Transactions on Pattern Analysis and Machine Intelligence},
  volume = {32},
  number = {8},
  pages = {1362--1376},
  year = {2010},
}

@inproceedings{galliani2015massively,
  author = {S. Galliani and K. Lasinger and K. Schindler},
  title = {{Massively parallel multiview stereopsis by surface normal diffusion}},
  booktitle = {ICCV},
  year = {2015},
}

@inproceedings{gropp2020igr,
  author = {A. Gropp and L. Yariv and N. Haim and M. Atzmon and Y. Lipman},
  title = {{Implicit geometric regularization for learning shapes}},
  booktitle = {ICML},
  year = {2020},
}

@article{hart1996sphere,
  author = {J. C. Hart},
  title = {{Sphere tracing: A geometric method for the antialiased ray tracing of implicit surfaces}},
  journal = {The Visual Computer},
  volume = {12},
  number = {10},
  pages = {527--545},
  year = {1996},
}

@inproceedings{huang2024_2dgs,
  author = {B. Huang and Z. Yu and A. Chen and A. Geiger and S. Gao},
  title = {{2D Gaussian splatting for geometrically accurate radiance fields}},
  booktitle = {ACM SIGGRAPH Conference Papers},
  year = {2024},
}

@inproceedings{jaderberg2015stn,
  author = {M. Jaderberg and K. Simonyan and A. Zisserman and K. Kavukcuoglu},
  title = {{Spatial transformer networks}},
  booktitle = {NeurIPS},
  year = {2015},
}

@inproceedings{jensen2014large,
  author = {R. Jensen and A. Dahl and G. Vogiatzis and E. Tola and H. Aan{\ae}s},
  title = {{Large scale multi-view stereopsis evaluation}},
  booktitle = {CVPR},
  year = {2014},
}

@inproceedings{jiang2023coordinate,
  author = {S. Jiang and J. Hua and Z. Han},
  title = {{Coordinate quantized neural implicit representations for multi-view reconstruction}},
  booktitle = {Proceedings of the IEEE/CVF International Conference on Computer Vision (ICCV)},
  year = {2023},
  pages = {18358--18369},
}

@inproceedings{kazhdan2006poisson,
  author = {M. Kazhdan and M. Bolitho and H. Hoppe},
  title = {{Poisson surface reconstruction}},
  booktitle = {Symposium on Geometry Processing},
  year = {2006},
  pages = {61--70},
}

@article{kerbl2023gaussian,
  author = {B. Kerbl and G. Kopanas and T. Leimk{\"u}hler and G. Drettakis},
  title = {{3D Gaussian Splatting for Real-Time Radiance Field Rendering}},
  journal = {ACM Transactions on Graphics},
  volume = {42},
  number = {4},
  pages = {139:1--139:14},
  year = {2023},
}

@misc{lenain2026surface,
  author = {Gabriel Lenain and Edmond Boyer and Jean Ponce},
  title = {{Surface reconstruction from range scans under visibility constraints}},
  howpublished = {arXiv preprint arXiv:2610.18005},
  year = {2026},
}

@inproceedings{li2026ambisur,
  author = {J. Li and J. Zhang and X. Bai and J. Zheng and X. Yu and L. Gu and G. H. Lee},
  title = {{Revisiting photometric ambiguity for accurate Gaussian-splatting surface reconstruction}},
  booktitle = {Proceedings of the 43rd International Conference on Machine Learning},
  year = {2026},
  pages = {72237--72259},
}

@inproceedings{li2023neuralangelo,
  author = {Z. Li and T. M{\"u}ller and A. Evans and R. H. Taylor and M. Unberath and M.-Y. Liu and C.-H. Lin},
  title = {{Neuralangelo: High-fidelity neural surface reconstruction}},
  booktitle = {CVPR},
  year = {2023},
}

@inproceedings{liu2020dist,
  author = {S. Liu and Y. Zhang and S. Peng and B. Shi and M. Pollefeys and Z. Cui},
  title = {{DIST: Rendering deep implicit signed distance function with differentiable sphere tracing}},
  booktitle = {CVPR},
  year = {2020},
}

@inproceedings{lorensen1987marching,
  author = {W. E. Lorensen and H. E. Cline},
  title = {{Marching cubes: A high resolution 3D surface construction algorithm}},
  booktitle = {SIGGRAPH},
  year = {1987},
}

@inproceedings{merrell2007realtime,
  author = {P. Merrell and A. Akbarzadeh and L. Wang and P. Mordohai and J.-M. Frahm and R. Yang and D. Nist{\'e}r and M. Pollefeys},
  title = {{Real-time visibility-based fusion of depth maps}},
  booktitle = {ICCV},
  year = {2007},
}

@inproceedings{meunier2022cpl,
  author = {L. Meunier and B. J. Delattre and A. Araujo and A. Allauzen},
  title = {{A dynamical system perspective for Lipschitz neural networks}},
  booktitle = {ICML},
  year = {2022},
  pages = {15484--15500},
}

@inproceedings{mildenhall2020nerf,
  author = {B. Mildenhall and P. P. Srinivasan and M. Tancik and J. T. Barron and R. Ramamoorthi and R. Ng},
  title = {{NeRF: Representing scenes as neural radiance fields for view synthesis}},
  booktitle = {ECCV},
  year = {2020},
}

@inproceedings{niemeyer2020dvr,
  author = {M. Niemeyer and L. Mescheder and M. Oechsle and A. Geiger},
  title = {{Differentiable volumetric rendering: Learning implicit 3D representations without 3D supervision}},
  booktitle = {CVPR},
  year = {2020},
}

@article{pons2007multi,
  author = {J.-P. Pons and R. Keriven and O. Faugeras},
  title = {{Multi-view stereo reconstruction and scene flow estimation with a global image-based matching score}},
  journal = {International Journal of Computer Vision},
  volume = {72},
  number = {2},
  pages = {179--193},
  year = {2007},
}

@inproceedings{prach2024layers,
  author = {B. Prach and F. Brau and G. Buttazzo and C. H. Lampert},
  title = {{1-Lipschitz layers compared: Memory, speed, and certifiable robustness}},
  booktitle = {CVPR},
  year = {2024},
  pages = {24574--24583},
}

@inproceedings{ren2024improving,
  author = {X. Ren and C. Cao and Y. Fu and X. Xue},
  title = {{Improving neural surface reconstruction with feature priors from multi-view images}},
  booktitle = {Proceedings of the European Conference on Computer Vision (ECCV)},
  year = {2024},
  pages = {445--463},
}

@inproceedings{schoenberger2016pixelwise,
  author = {J. L. Sch{\"o}nberger and E. Zheng and J.-M. Frahm and M. Pollefeys},
  title = {{Pixelwise view selection for unstructured multi-view stereo}},
  booktitle = {European Conference on Computer Vision (ECCV)},
  year = {2016},
  pages = {501--518},
}

@inproceedings{tancik2020fourier,
  author = {M. Tancik and P. Srinivasan and B. Mildenhall and S. Fridovich-Keil and N. Raghavan and U. Singhal and R. Ramamoorthi and J. Barron and R. Ng},
  title = {{Fourier features let networks learn high frequency functions in low dimensional domains}},
  booktitle = {NeurIPS},
  year = {2020},
}

@inproceedings{toussaint2025probesdf,
  author = {B. Toussaint and D. Thomas and J.-S. Franco},
  title = {{ProbeSDF: Light Field Probes for Neural Surface Reconstruction}},
  booktitle = {CVPR},
  year = {2025},
}

@article{vicini2022differentiable,
  author = {D. Vicini and S. Speierer and W. Jakob},
  title = {{Differentiable signed distance function rendering}},
  journal = {ACM Transactions on Graphics},
  volume = {41},
  number = {4},
  pages = {125:1--125:18},
  year = {2022},
}

@article{vu2012highaccuracy,
  author = {H.-H. Vu and P. Labatut and J.-P. Pons and R. Keriven},
  title = {{High accuracy and visibility-consistent dense multiview stereo}},
  journal = {IEEE Transactions on Pattern Analysis and Machine Intelligence},
  volume = {34},
  number = {5},
  pages = {889--901},
  year = {2012},
}

@inproceedings{wang2021neus,
  author = {P. Wang and L. Liu and Y. Liu and C. Theobalt and T. Komura and W. Wang},
  title = {{NeuS: Learning neural implicit surfaces by volume rendering for multi-view reconstruction}},
  booktitle = {NeurIPS},
  year = {2021},
}

@inproceedings{wang2026surfaces,
  author = {S. Wang and D. J. Miller},
  title = {{When 3D Gaussian Splatting Recovers Real Surfaces}},
  booktitle = {European Conference on Computer Vision (ECCV)},
  year = {2026},
  pages = {621--639},
}

@inproceedings{weidenfeller2026adele,
  author = {J. Weidenfeller and S. Wang and P. F{\"u}rnstahl and S. Tang},
  title = {{ADELE: Adaptive Delaunay grids for high-fidelity mesh-native reconstruction}},
  booktitle = {SIGGRAPH Asia Conference Papers},
  year = {2026},
}

@article{xu2022acmmp,
  author = {Q. Xu and W. Kong and W. Tao and M. Pollefeys},
  title = {{Multi-scale geometric consistency guided and planar prior assisted multi-view stereo}},
  journal = {IEEE Transactions on Pattern Analysis and Machine Intelligence},
  volume = {45},
  number = {4},
  pages = {4945--4963},
  year = {2023},
}

@inproceedings{yang2023steik,
  author = {H. Yang and Y. Sun and G. Sundaramoorthi and A. Yezzi},
  title = {{StEik: Stabilizing the optimization of neural signed distance functions and finer shape representation}},
  booktitle = {NeurIPS},
  year = {2023},
}

@inproceedings{yao2020blendedmvs,
  author = {Y. Yao and Z. Luo and S. Li and J. Zhang and Y. Ren and L. Zhou and T. Fang and L. Quan},
  title = {{BlendedMVS: A large-scale dataset for generalized multi-view stereo networks}},
  booktitle = {CVPR},
  year = {2020},
}

@inproceedings{yariv2020idr,
  author = {L. Yariv and Y. Kasten and D. Moran and M. Galun and M. Atzmon and R. Basri and Y. Lipman},
  title = {{Multiview neural surface reconstruction by disentangling geometry and appearance}},
  booktitle = {NeurIPS},
  year = {2020},
}

@inproceedings{yariv2021volsdf,
  author = {L. Yariv and J. Gu and Y. Kasten and Y. Lipman},
  title = {{Volume rendering of neural implicit surfaces}},
  booktitle = {NeurIPS},
  year = {2021},
}

@inproceedings{zhang2021mvsdf,
  author = {J. Zhang and Y. Yao and L. Quan},
  title = {{Learning signed distance field for multi-view surface reconstruction}},
  booktitle = {ICCV},
  year = {2021},
  pages = {6525--6534},
}

@misc{zhang2020nerfpp,
  author = {K. Zhang and G. Riegler and N. Snavely and V. Koltun},
  title = {{NeRF++: Analyzing and improving neural radiance fields}},
  howpublished = {arXiv preprint arXiv:2010.07492},
  year = {2020},
}
}

\maketitlesupplementary
\setcounter{figure}{7}
\setcounter{table}{4}
\setcounter{equation}{10}
\appendix
\section{Additional experimental details}

\subsection{Experimental parameters. }

\paragraph{Scene normalization.} Following IDR~\cite{yariv2020idr} and
NeuS~\cite{wang2021neus}, we center and scale each scene to approximately unit sphere coordinates using the provided camera normalization. All spatial hyperparameters are expressed in these coordinates.

\paragraph{Sphere-tracing parameters.}
\label{app:st-params}
We trace surface intersections with at most 128 iterations and tolerance $10^{-4}$, and visibility queries with at most 64 iterations and tolerance $10^{-3}$. $10^{-4}$ corresponds to
approximately $0.02$ mm on DTU.
\\
For reproducibility, Table~\ref{tab:hyperparameters}
reports the remaining implementation parameters and their values.
\begin{table}[H]
\centering
\caption{Additional implementation parameters.}
\label{tab:hyperparameters}
\small
\setlength{\tabcolsep}{6pt}
\begin{tabular}{lll}
\toprule
\textbf{Parameter} & \textbf{Description} & \textbf{Value} \\
\midrule
$T$ & Maximum tracing distance & $5$ \\
$\Delta$ & Occlusion tolerance & $10^{-2}$ \\
$\kappa$ & Grazing angle threshold & $0.1$ \\
$\tau$ & Photometric threshold & $0$ \\
$\rho$ & Patch spacing & $1$\,px \\
\midrule
$\lambda$ & Depth confidence threshold & $0.5$ \\
$\mu$ & Carving depth margin & $3$ voxels \\
$\nu$ & Minimum carving votes & $3$ \\
\bottomrule
\end{tabular}
\end{table}
\noindent
Even when tracing starts at the bounding sphere, \(T\) is measured from the camera center.

\subsection{Initialization}
\label{app:init}
We first construct a coarse geometric proxy.
\paragraph{Depth-confidence carving.}
For each scene, we use ACMMP~\cite{xu2022acmmp} to compute depth maps
\(D_i\in\mathbb R_+^{H_i\times W_i}\) and confidence maps
\(C_i\in[0,1]^{H_i\times W_i}\).
Starting from an enclosing sphere \(\Omega_0\), we remove voxels
that are confidently observed in free space. Formally, for a voxel
\(x\in\Omega_0\), let
\begin{equation}
b_i(x)=\mathds{1}\!\big[
C_i(\pi_i(x))\geq\lambda,\;
z_i(x)+\mu h<D_i(\pi_i(x))
\big],
\end{equation}
where \(h\) is the voxel spacing, \(\lambda\in[0,1]\) is the minimum confidence threshold and \(\mu>0\) sets the carving margin in voxels. We retain
\begin{equation}
\Omega^\star
=
\left\{
x\in\Omega_0 \;\middle|\;
\sum_{i=1}^n b_i(x)<\nu
\right\}.
\end{equation}
Thus, a voxel is carved only when sufficiently many confident views agree
that it lies in front of the observed surface.

\paragraph{Continuous-field initialization.}
We then compute a discrete SDF \(\phi\)
from $\Omega^*$. For every voxel $k$ we have 
\begin{equation}
\phi(k) = h\left[
d(k,\Omega^\star) -
d(k,\Omega_0\setminus\Omega^\star)
\right],
\end{equation}
where $d(k,A)=\min_{\ell\in A}\|k-\ell\|_2$.
 Values between voxel centers are obtained by trilinear interpolation and we initialize by fitting
\begin{equation}
\theta_0
=
\arg\min_\theta
\mathbb E_{x\sim p_0}
\left[
(f_\theta(x)-\phi(x))^2
\right],
\end{equation}
where \(p_0\) mixes uniform samples with samples concentrated near
\(\partial\Omega^\star\).
\paragraph{Why this initialization? }
\begin{itemize}
    \item It is closed by construction, providing a valid signed target, unlike trimmed Poisson meshes.
    \item It is conservative, the proxies carved from ACMMP and COLMAP depth maps enclose $98.8\%$ and
$99.4\%$ of DTU ground-truth points, respectively, providing
conservative initializations.
    \item It is coarse (1.55\,mm), uses only
the input images, no masks or pretrained priors, and takes about $10$
minutes per scene.
\end{itemize}

\subsection{Orthonormal basis of $n_\theta(x)^\perp$}
\label{sec:orthonormal-basis}
To define the patch at $x$, we construct an orthonormal basis
$(e_1(x),e_2(x),n_\theta(x))$ of $\mathbb{R}^3$. Let
\[
w(n_\theta(x))
=
\begin{cases}
(0,1,0)^\top, & \text{if } |n_\theta(x)^\top(0,1,0)^\top|<0.9,\\
(1,0,0)^\top, & \text{otherwise}.
\end{cases}
\]
We then define
\begin{equation}
\label{eq:tangent-frame}
\left\{
\begin{aligned}
e_1(x)
&=
\frac{
n_\theta(x)\times w(n_\theta(x))
}{
\|n_\theta(x)\times w(n_\theta(x))\|_2
},
\\[0.2em]
e_2(x)
&=
n_\theta(x)\times e_1(x).
\end{aligned}
\right.
\end{equation}

\subsection{Evaluation metrics}
\label{sec:eval-metrics}

\paragraph{DTU.}
We follow the standard DTU evaluation protocol~\cite{jensen2014large}.
Let $P$ be the point cloud sampled from the reconstructed mesh and $G$
the official ground-truth point cloud. We report the symmetric
Chamfer-$L_1$ distance, in millimeters:
\begin{equation}
\mathrm{CD_1}(P,G)
=
\frac{1}{2}
\left(
\frac{1}{|P|}\sum_{p\in P} d(p,G)
+
\frac{1}{|G|}\sum_{g\in G} d(g,P)
\right),
\end{equation}
where $d(x,Q)=\min_{q\in Q}\|x-q\|_2$. Predicted points are
filtered by the observation mask dilated by 12 pixels, ground-truth points by
the ground plane and distances $\geq 20$ mm are excluded \cite{yariv2020idr}.

\paragraph{Visibility culling.}
\label{sec:visibility-cull}

\begin{figure}[H]
    \centering
    \includegraphics[width=\columnwidth]{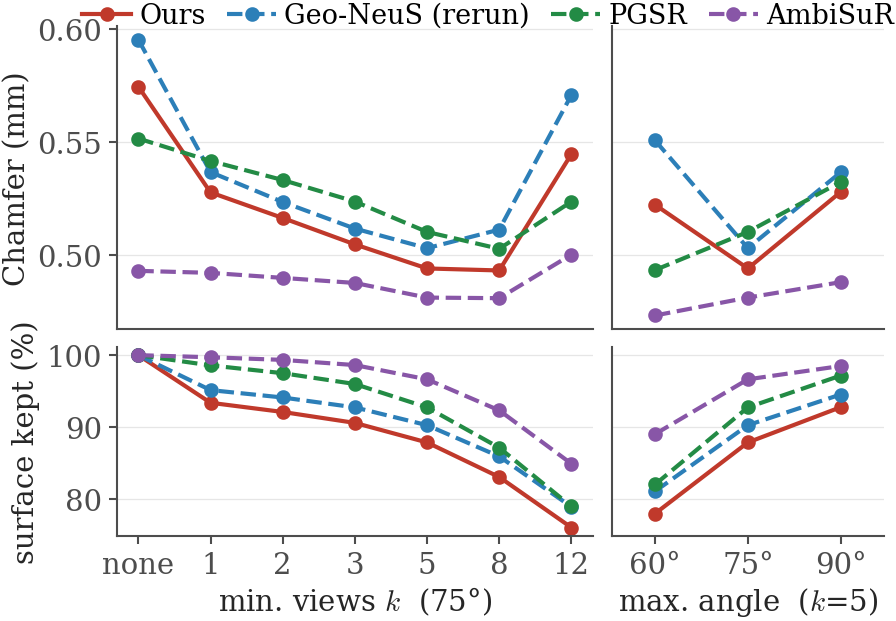}
    \caption{
    Effect of visibility culling thresholds on DTU. Mean $\mathrm{CD}_1$ (top) and fraction of mesh area retained (bottom) as functions of the minimum number of supporting views $k$ and maximum viewing angle $\alpha$.}
    \label{fig:visibility_culling}
\end{figure}

We apply the same visibility culling to the meshes of all methods after their respective postprocessing.  Let $\mathcal{M}=(\mathcal{V},\mathcal{F})$ be a mesh with
face centroids $\mathbf{m}_f$ and unit normals $\mathbf{n}_f$.
For each training camera $i$, we raycast all
pixels at full resolution to obtain first intersection distances $d_i(p)$.
Writing $p_i=\lfloor\pi_i(\mathbf{m}_f)\rfloor$ and
$\mathbf{u}_i=(\mathbf{m}_f-\mathbf{c}_i)/
\|\mathbf{m}_f-\mathbf{c}_i\|_2$,
a camera supports face $f$ if $p_i$ is valid,
$d_i(p_i)$ is finite and
\begin{equation}
\|\mathbf{m}_f-\mathbf{c}_i\|_2
\leq (1+\eta)d_i(p_i),
\qquad
|\mathbf{n}_f^\top\mathbf{u}_i|\geq\cos\alpha,
\end{equation}
where $\eta=2\times10^{-3}$ accounts for numerical tolerance.
Denoting the supporting views by $V_\alpha(f)$, we retain
\begin{equation}
\mathcal{F}_{k,\alpha}
=\{f\in\mathcal{F}:|V_\alpha(f)|\geq k\}.
\end{equation}
We use $(k,\alpha)=(5,75^\circ)$ for DTU. Figure~\ref{fig:visibility_culling} reports sensitivity to these thresholds, while Figure~\ref{fig:culling_visualization} shows the removed
geometry across three DTU scans.

\paragraph{BMVS.}
We follow the BMVS evaluation protocol of ProbeSDF~\cite{toussaint2025probesdf}, changing only the surface sampling: points are sampled uniformly by area rather than at mesh vertices to reduce dependence on tessellation. We then compute the bidirectional unsigned point-to-surface distance between the reconstruction and ground-truth meshes, discard distances larger than $0.025$ in normalized scene coordinates, and report the mean after multiplication by $10^3$.

\subsection{Architectural details}
\label{sec:architecture-details}
We summarize the main components of the CPL--MaxMin
architecture~\citep{anil2019groupsort,meunier2022cpl,prach2024layers}
for completeness.

\paragraph{Input embedding.}
Following~\citet{meunier2022cpl}, we embed the input into
$\mathbb{R}^D$ using fixed zero-padding:
\begin{equation}
A=\begin{bmatrix}I_{m_{\text{in}}}\\0\end{bmatrix},
\qquad
m_{\text{in}}=\begin{cases}
3 & \text{without PE},\\
6L+3 & \text{with PE}.
\end{cases}
\end{equation}
This map preserves Euclidean distances since
$A^\top A=I_{m_{\text{in}}}$.

\paragraph{MaxMin activation.}
The backbone $\Phi_\theta$ alternates CPL layers with
MaxMin activations, which sort successive pairs of coordinates.
For $z\in\mathbb{R}^D$,
\begin{equation}
\operatorname{MaxMin}(z)
=
\bigl(
\max(z_{2k-1},z_{2k}),
\min(z_{2k-1},z_{2k})
\bigr)_{k=1}^{D/2}.
\end{equation}
Pairwise sorting preserves the Euclidean norm and is
$1$-Lipschitz~\citep{anil2019groupsort}.

\paragraph{Convex Potential Layers.}
Following~\citet{meunier2022cpl}, each CPL is a residual
mapping of the form
\begin{equation}
\ell(z)
=
z-
\frac{2}{\|W\|_2^2}
W^\top\sigma(Wz+b),
\end{equation}
where $\sigma=\operatorname{ReLU}$,
$W\in\mathbb{R}^{m\times D}$, and $\|W\|_2$
denotes the spectral norm.
This construction is $1$-Lipschitz when the exact
spectral norm is used.

\paragraph{Spectral norm estimation.}
In practice, we approximate $\|W\|_2$ by power
iteration~\citep{meunier2022cpl}.
Starting from a unit vector $u$, we alternate
normalized multiplications by $W$ and $W^\top$:
\begin{equation}
v
\leftarrow
\frac{Wu}{\|Wu\|_2},
\qquad
u
\leftarrow
\frac{W^\top v}{\|W^\top v\|_2}.
\end{equation}
The spectral norm is then estimated as
\begin{equation}
\widehat{\sigma}(W)=v^\top Wu.
\end{equation}
We perform one power-iteration update per optimization step.

\subsection{Baselines}

\paragraph{Gaussian-splatting baselines.}
Since default Gaussian splatting configurations yield reconstruction failures and large errors on BMVS, we rescale the position learning rate and densification threshold of 2DGS, PGSR, and AmbiSuR by $R_0/R$, where $R=1.1\max_i||c_i-\bar c||_2$ and $R_0=2.157$ is the mean DTU camera extent. This generally improves robustness and mean accuracy, although not uniformly across scenes (Table~\ref{tab:gs-rescaling}).

\begin{table}[H]
\centering
\caption{Effect of camera extent normalization on BMVS.
Chamfer-$L_1$ ($\times 10^{3}$).
Best results are in bold. Fail indicates reconstruction failure.}
\label{tab:gs-rescaling}
\scriptsize
\setlength{\tabcolsep}{3pt}
\renewcommand{\arraystretch}{1.05}
\resizebox{\columnwidth}{!}{%
\begin{tabular}{lcccccc}
\toprule
& \multicolumn{2}{c}{2DGS}
& \multicolumn{2}{c}{PGSR}
& \multicolumn{2}{c}{AmbiSuR} \\
\cmidrule(lr){2-3}
\cmidrule(lr){4-5}
\cmidrule(lr){6-7}
Scene & Default & Rescaled & Default & Rescaled
& Default & Rescaled \\
\midrule
Dog       & 6.50 & \textbf{4.61} & \textbf{2.15} & 2.29 & 1.58 & \textbf{1.57} \\
Bear      & \textbf{11.54} & 12.32 & 1.54 & \textbf{1.51} & 1.30 & \textbf{1.29} \\
Clock     & \textbf{8.64} & 11.03 & 3.57 & \textbf{2.92} & 11.09 & \textbf{3.21} \\
Durian    & 8.76 & \textbf{6.60} & 2.69 & \textbf{2.66} & 2.74 & \textbf{2.68} \\
Man       & 5.83 & \textbf{4.37} & \textbf{1.81} & 1.96 & \textbf{1.64} & 1.68 \\
Sculpture & \textbf{11.59} & 12.39 & 2.08 & \textbf{2.04} & \textbf{1.84} & 1.89 \\
Stone     & 5.15 & \textbf{4.32} & 2.01 & \textbf{1.98} & \textbf{1.88} & 1.93 \\
Jade      & \textbf{4.77} & 4.92 & Fail & \textbf{5.10} & 8.54 & \textbf{5.27} \\
\bottomrule
\end{tabular}%
}
\end{table}

\paragraph{Neuralangelo.}
\label{app:neuralangelo_bmvs}
Using the official implementation, we evaluate Neuralangelo on all 15 DTU scans and eight BMVS scenes (Table~\ref{tab:neuralangelo_results}). On DTU, we obtain a mean $\mathrm{CD}_1$ of 1.71 mm versus the published 0.61 mm, Ren et al.~\cite{ren2024improving} independently report 2.01 mm. On BMVS, using the DTU configuration, four scenes exhibit substantially higher errors, accompanied by low sigmoid sharpness values $s$ (28--55 versus 257--534 on the remaining scenes). Given these reproduction difficulties, we exclude Neuralangelo from the main comparisons.

\begin{table}[H]
\centering
\caption{Neuralangelo reproduction on BMVS and DTU.
DTU results use the standard 12-pixel silhouette crop.}
\label{tab:neuralangelo_results}
\small
\begin{tabular}{lc@{\hspace{1.2cm}}lc}
\toprule
\multicolumn{2}{c}{BMVS ($\mathrm{CD}_1$, $\times 10^{3}$)} &
\multicolumn{2}{c}{DTU ($\mathrm{CD}_1$, mm)} \\
\cmidrule(lr){1-2}\cmidrule(lr){3-4}
Scene & Error & Scan & Error \\
\midrule
Dog       & 12.21 & 24  & 0.40 \\
Bear      & 2.20  & 37  & 0.76 \\
Clock     & 1.79  & 40  & 0.56 \\
Durian    & 3.46  & 55  & 0.39 \\
Man       & 9.58  & 63  & 1.90 \\
Sculpture & 11.13 & 65  & 0.53 \\
Stone     & 1.77  & 69  & 2.54 \\
Jade      & 9.63  & 83  & 2.04 \\
          &       & 97  & 3.19 \\
          &       & 105 & 2.33 \\
          &       & 106 & 1.21 \\
          &       & 110 & 2.27 \\
          &       & 114 & 0.65 \\
          &       & 118 & 4.29 \\
          &       & 122 & 2.65 \\
\midrule
Mean & 6.47 & Mean & 1.71 \\
\bottomrule
\end{tabular}
\end{table}

\section{Computation and sphere tracing}

\subsection{Lipschitz preliminaries}
\label{app:lipschitz-def}

For \(f:\mathbb{R}^d\to\mathbb{R}^m\), define
\begin{equation}
\Lip(f)
=
\sup_{x\neq x'}
\frac{\|f(x)-f(x')\|_2}{\|x-x'\|_2}.
\end{equation}
The function \(f\) is Lipschitz if \(\Lip(f)<\infty\), and
\(C\)-Lipschitz, for \(C\ge0\), if \(\Lip(f)\le C\).

\subsection{Proofs}
\label{app:proofs}

\paragraph{Sphere-tracing convergence.}
We recall the standard convergence argument~\citep{hart1996sphere}
for our camera rays.
Let $r(t)=c_i+tu_i(p)$ and
$t^\star=t_\theta(i,p)<\infty$ denote the first zero
of $f_\theta\circ r$, with $\|u_i(p)\|_2=1$.
By the $1$-Lipschitz property, for any $t_k\leq t^\star$,
\begin{equation}
0\leq t_{k+1}-t_k
=|f_\theta(r(t_k))|
\leq t^\star-t_k.
\end{equation}
Starting from $t_0=0$, the sequence is therefore
nondecreasing and bounded by $t^\star$.
At its limit $\bar t$, the increments vanish, so
continuity gives $f_\theta(r(\bar t))=0$.
Since $t^\star$ is the first zero, $\bar t=t^\star$.
\\
If $f_\theta\circ r$ is $C^1$ near $t^\star$
and the intersection is non-grazing,
its derivative is bounded away from zero locally.
The mean value theorem then gives
$|f_\theta(r(t))|\geq c(t^\star-t)$
for some $c\in(0,1]$ and all $t<t^\star$
sufficiently close to $t^\star$.
Consequently, the error $e_k=t^\star-t_k$ satisfies
\begin{equation}
e_{k+1}\leq(1-c)e_k,
\end{equation}
establishing local linear convergence.
\hfill$\square$

\subsection{Efficient sphere tracing}
\label{sec:efficient-st}
\begin{enumerate}[label=(\roman*)]
    \item After each iteration, rays that have converged or left the valid depth interval are excluded from subsequent network evaluations.
    \item The final non-grazing intersection is differentiated implicitly using Eq~\ref{eq:intersection-gradient}, rather than backpropagating through all
tracing iterations.
    \item Following DIST~\cite{liu2020dist}, sphere tracing starts at the intersection
between the camera ray and the enclosing sphere $\Omega_0$, preventing useless evaluations of
\(f_\theta\) in empty space.
\end{enumerate}

\subsection{Computational cost}
\label{sec:computational}

\paragraph{Normal differentiation.}
Our implementation stops gradients through the tangent plane normals,
avoiding second-order derivatives.
Differentiating through the normals increases iteration time by 20\%
and GPU memory by 140\%, while worsening mean $\mathrm{CD}_1$
on five DTU scans (0.525 vs.\ 0.478 mm).

\begin{table}[!ht]
\centering
\caption{Approximate training time per DTU scene, all measured on the same NVIDIA H100 GPU.}
\label{tab:runtime}
\small
\setlength{\tabcolsep}{6pt}
\begin{tabular}{lc}
\toprule
\textbf{Method} & \textbf{Time $\downarrow$} \\
\midrule
2DGS         & $\sim$10 min \\
PGSR         & $\sim$20 min \\
AmbiSuR      & $\sim$20 min \\
\midrule
NeuS         & $\sim$4 h \\
NeuralWarp   & $\sim$4 h \\
Geo-NeuS     & $\sim$5 h \\
Neuralangelo & $\sim$7 h \\
\midrule
\textbf{Ours (total)} & $\boldsymbol{\sim}$\textbf{5 h 10 min} \\
\quad ACMMP initialization & $\sim$10 min \\
\quad Photometric refinement & $\sim$5 h \\
\bottomrule
\end{tabular}
\end{table}

\subsection{Sphere tracing reliability}

For each model, we uniformly sample image rays until 100k
yield a sphere-tracing intersection ($M=128$).
Using 192-bit Arb interval arithmetic, we certify the absence
of earlier roots and enclose the first intersection within
$10^{-7}r_0$, where $r_0$ is the radius of the enclosing sphere
$\Omega_0$. On rays with certified intersections, we evaluate
first-hit accuracy and overshoots with a tolerance of
$10^{-3}r_0$. Gradient statistics use the maximum sampled
$\|\nabla_x f_\theta\|_2$ along each ray.

\begin{table}[t]
\centering
\caption{Sphere tracing reliability on DTU.}
\label{tab:tracing-ablation}
\footnotesize
\setlength{\tabcolsep}{3pt}
\renewcommand{\arraystretch}{1.05}
\begin{tabular}{@{}lccccc@{}}
\toprule
\textbf{PE}
& \textbf{Overshoots}
& \textbf{Hit (\%)}
& \textbf{Q99}
& \textbf{Max}
& \textbf{$>1$ (\%)} \\
\midrule
$L=0$ & 0 & 98.38 & 0.962 & 0.982 & 0.000 \\
$L=6$ & 0 & 99.83 & 0.873 & 1.106 & 0.019 \\
\bottomrule
\end{tabular}
\end{table}

\section{Additional results and analyses}

\subsection{Photometric localization}
\begin{figure*}[t]
    \centering
    \includegraphics[width=\textwidth]{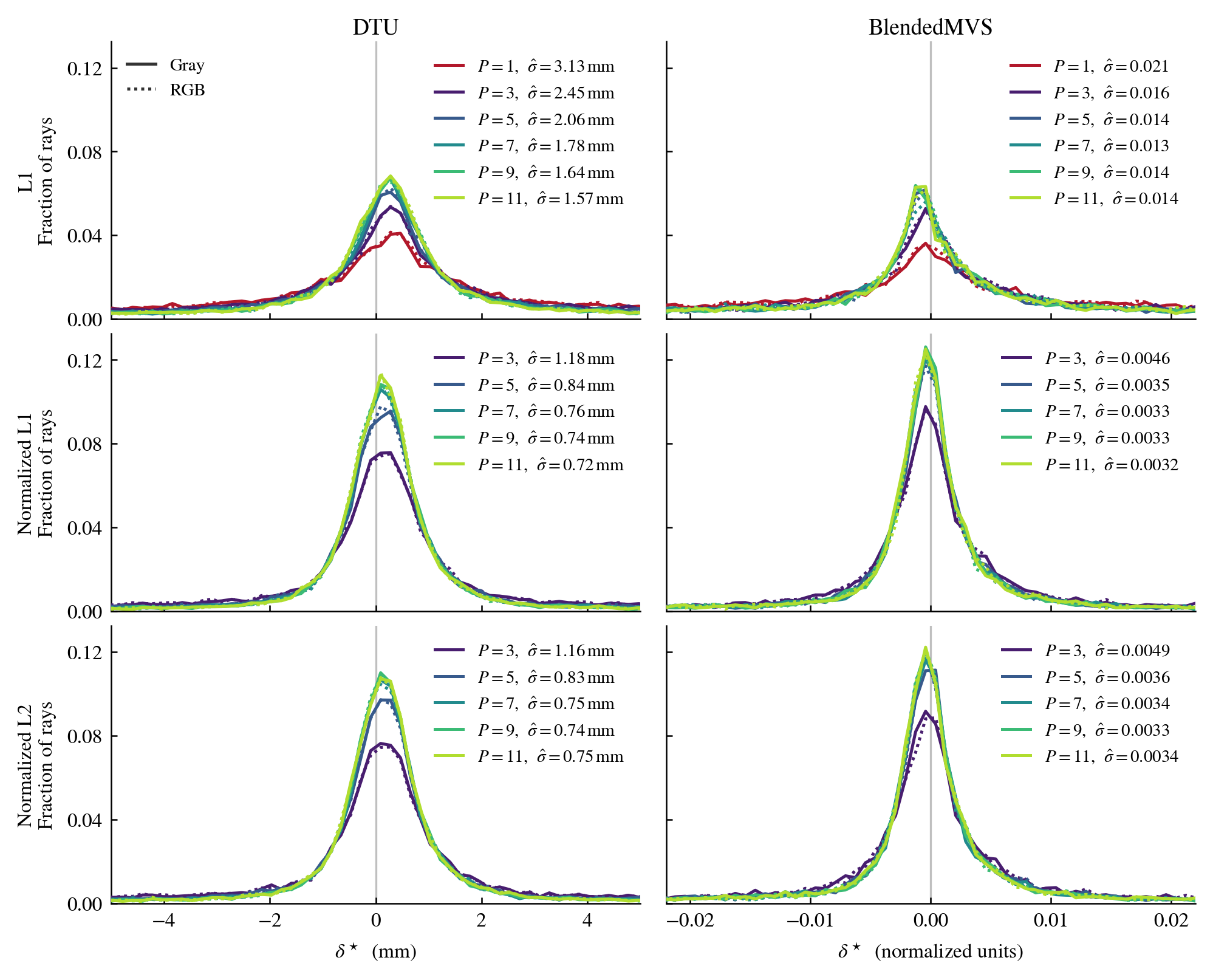}
    \caption{
Photometric depth localization on DTU and BMVS.
Distributions of depth errors $\delta^\star=\hat t-t^\star$, where $\hat t$ minimizes Eq.~\eqref{eq:localization},
for raw patch $\ell_1$ (top), normalized patch $\ell_1$ (middle), and normalized patch squared $\ell_2$ (bottom),
with patch sizes $P$, using grayscale (solid) and RGB (dotted).
$\hat\sigma$ denotes the estimated standard deviation of $\delta^\star$.
}
    \label{fig:photometric_localization}
\end{figure*}

To evaluate how accurately different photometric losses recover ground truth depth, we estimate the depth of each reference ray by minimizing
\begin{equation}
\hat t=\arg\min_{ t\in\mathcal T}\frac{1}{|\mathcal N|}\sum_{j\in\mathcal N}d(A_i(t),A_j(t)),
\label{eq:localization}
\end{equation}
where $\mathcal T$ contains sampled candidate depths and $d$ denotes raw patch $\ell_1$, normalized patch $\ell_1$, or normalized patch squared $\ell_2$ (Fig.~\ref{fig:photometric_localization}, top to bottom). $\mathcal N$ contains the $K=6$ nearest source views,
held fixed across candidate depths. On DTU and BMVS, both normalized losses yield comparable accuracy and substantially outperform raw $\ell_1$. Larger patches improve localization, with diminishing gains beyond $P=7$, while grayscale and RGB perform similarly.

\subsection{Geometry--appearance ambiguity}

\paragraph{Volumetric rendering.}
Let $\theta$ and $\varphi$ denote geometry and appearance
parameters, where $\varphi$ may be
color MLP weights~\cite{wang2021neus} or spherical
harmonic coefficients~\cite{kerbl2023gaussian}.
Following Geo-NeuS~\cite{fu2022geoneus}, let
$\hat C(\theta,\varphi)$ denote the rendered color along a ray.
At a solution matching the observed colors,
a first-order perturbation gives
\begin{equation}
\delta\hat C
=
J_\theta\delta\theta+J_\varphi\delta\varphi,
\end{equation}
where $J_\theta=\partial\hat C/\partial\theta$ and
$J_\varphi=\partial\hat C/\partial\varphi$.
Optimizing the appearance perturbation yields
\begin{equation}
\min_{\delta\varphi}\|\delta\hat C\|_2^2
=
\|(I-J_\varphi J_\varphi^\dagger)
J_\theta\delta\theta\|_2^2.
\end{equation}
Consequently, geometric perturbations satisfying
$J_\theta\delta\theta\in\operatorname{Im}(J_\varphi)$
can be fully compensated by appearance changes to
first order. This local ambiguity is consistent with prior results showing
that expressive view-dependent appearance can reproduce
observed images despite incorrect geometry~\cite{zhang2020nerfpp,wang2026surfaces}.

\begin{figure}[H]
    \centering
    \includegraphics[width=\linewidth]{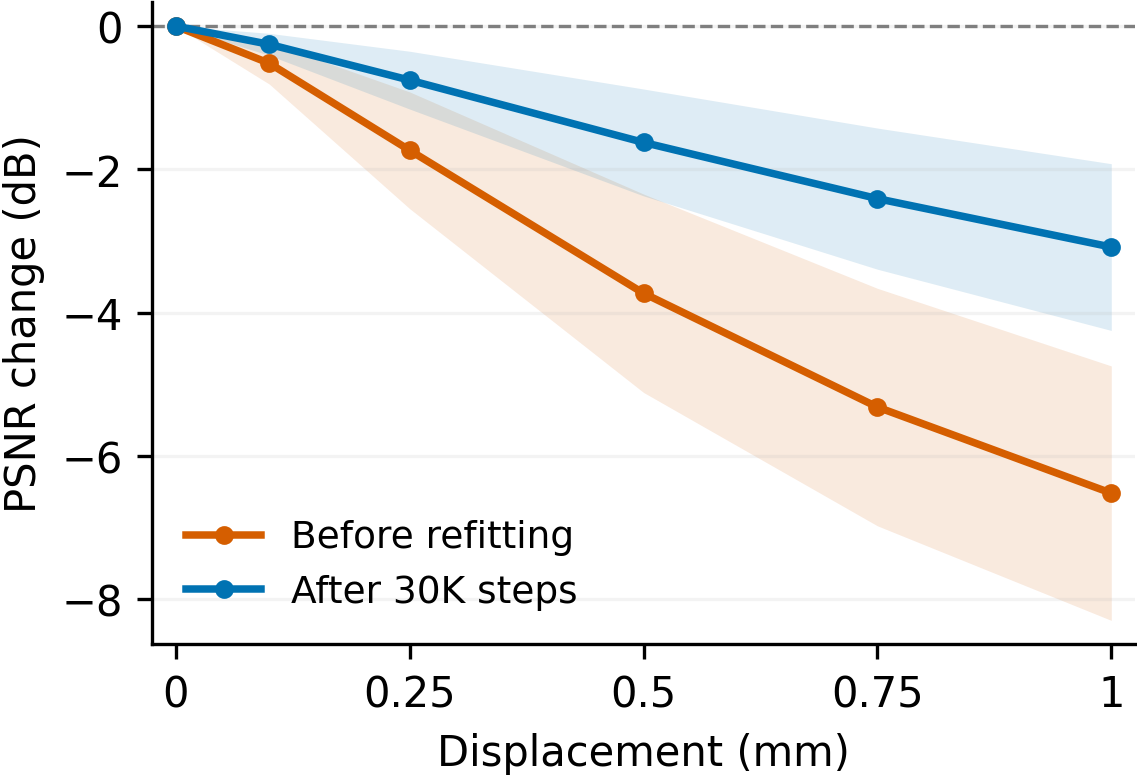}
    \caption{ Refitting only the appearance coefficients partially recovers PSNR
    despite fixed geometry. Shaded regions indicate variability across scenes.}
    \label{fig:appearance_compensation}
\end{figure}

\paragraph{Experimental illustration.}
To illustrate this, we displace the 3D Gaussian centers of AmbiSuR models trained on DTU by varying magnitudes $\delta \in \{0.10, 0.25, 0.5, 0.75, 1\}$mm. For each perturbation, we freeze all parameters except the spherical harmonics coefficients, which we optimize for 30k steps, including at $\delta=0$ as a control. Figure~\ref{fig:appearance_compensation} reports the PSNR degradation before and after refitting. Appearance optimization partially recovers rendering quality despite fixed geometric perturbations.

\paragraph{Direct patch photoconsistency.}
Our formulation has no appearance parameters $\varphi$.
Let $t$ denote the first-hit depth along a reference ray
and let $\tilde A_i(t)$ and $\tilde A_j(t)$ be the
zero-mean, unit-normalized observed patches from
Sec.~\ref{sec:photo-supervision}. We consider the equivalent loss
\begin{equation}
\mathcal L_{\mathrm{patch}}(t)
=
\frac{1}{4|\mathcal N|}
\sum_{j\in \mathcal N}
\|\tilde A_i(t)-\tilde A_j(t)\|_2^2,
\end{equation}
where $\mathcal{N}$ is the fixed set of accepted source views. Let $t^\star$ denote the ground truth depth. Assuming $\mathcal L_{\mathrm{patch}}(t^\star)=0$ and fixed patch orientation, we have
\begin{equation}
\mathcal L_{\mathrm{patch}}(t^\star+\delta t)
=
\frac{(\delta t)^2}{4|\mathcal N|}
\sum_{j\in \mathcal N}
\|\tilde A_j'(t^\star)\|_2^2
+o((\delta t)^2).
\end{equation}
Thus, the true depth is a strict local minimizer
whenever at least one source patch varies to first
order with depth. This is consistent with
Fig.~\ref{fig:photometric_localization}, where photometric
depth estimates concentrate around the ground truth
on both DTU and BMVS.

\subsection{Initialization}

\begin{figure}[H]
    \centering
    \includegraphics[width=\linewidth]{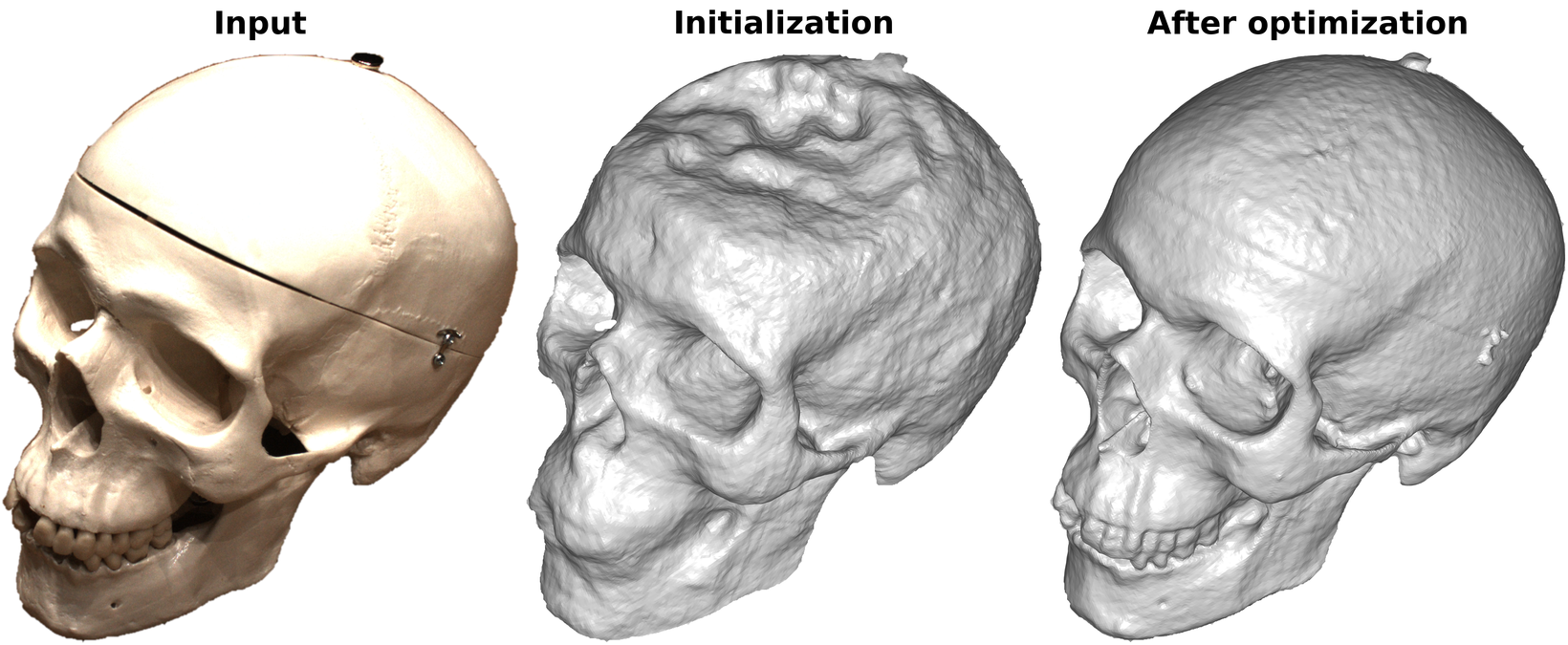}
    \caption{On DTU scan 65, photometric optimization corrects substantial geometric errors in the ACMMP initialization, reducing $\mathrm{CD}_1$ from 1.69 to 0.55 mm.}
    \label{fig:init_recovery}
\end{figure}

Table~\ref{tab:init_ablation} provides detailed results across all 15 DTU scans, while Figure~\ref{fig:init_recovery} illustrates how photometric refinement can correct substantial geometric errors from the ACMMP initialization.

\begin{table*}[!t]
\centering
\caption{Effect of initialization on DTU ($\mathrm{CD}_1$ in mm; Mean: mean $\pm$ standard deviation over the 15 scans).}
\label{tab:init_ablation}
\scriptsize
\setlength{\tabcolsep}{3pt}
\renewcommand{\arraystretch}{1.1}
\resizebox{\textwidth}{!}{%
\begin{tabular}{l*{16}{c}}
\toprule
\textbf{Initialization}
& \textbf{24} & \textbf{37} & \textbf{40}
& \textbf{55} & \textbf{63} & \textbf{65}
& \textbf{69} & \textbf{83} & \textbf{97}
& \textbf{105} & \textbf{106} & \textbf{110}
& \textbf{114} & \textbf{118} & \textbf{122}
& \textbf{Mean} \\
\midrule

\multicolumn{17}{l}{\textit{Coarse initialization}} \\

COLMAP
& 1.24 & 2.51 & 1.46 & 0.88 & 1.85
& 1.57 & 1.44 & 2.01 & 2.43 & 1.65
& 1.25 & 1.90 & 1.41 & 1.35 & 1.06
& 1.60\sd{0.47} \\ 

ACMMP
& 1.20 & 2.16 & 1.36 & 0.86 & 1.77
& 1.69 & 1.33 & 2.01 & 2.38 & 1.60
& 1.15 & 1.84 & 1.41 & 1.38 & 1.05
& 1.55\sd{0.43} \\ 

\midrule
\multicolumn{17}{l}{\textit{Refined}} \\
Sphere init
& 1.36 & 1.91 & 1.17 & 0.42 & 1.05
& 0.87 & 0.64 & \textbf{1.16} & 0.97 & 0.68
& 0.64 & \textbf{0.58} & 0.31 & 0.49 & 0.47
& 0.85\sd{0.43} \\ 

COLMAP init
& 0.47 & 0.83 & \textbf{0.38} & \textbf{0.35} & 0.85
& 0.59 & 0.52 & 1.26 & \textbf{0.90} & 0.65
& 0.56 & 0.75 & 0.31 & 0.40 & 0.40
& 0.61\sd{0.26} \\ 

\textbf{Ours}
& \textbf{0.41} & \textbf{0.74} & \textbf{0.38}
& \textbf{0.35} & \textbf{0.80} & \textbf{0.55}
& \textbf{0.48} & 1.20 & 0.91
& \textbf{0.62} & \textbf{0.46} & 0.62
& \textbf{0.30} & \textbf{0.39} & \textbf{0.37}
& \textbf{0.57}\sd{0.25} \\

\bottomrule
\end{tabular}%
}
\end{table*}

\paragraph{Classical mesh refinement.}
Table~\ref{tab:classical-refinement} compares classical photometric mesh refinement~\citep{vu2012highaccuracy} with our method. 

\begin{table}[H]
\centering
\caption{Photometric refinement on DTU (mean $\mathrm{CD}_1$ in mm).}
\label{tab:classical-refinement}
\small
\setlength{\tabcolsep}{6pt}
\begin{tabular}{llcc}
\toprule
\textbf{Method} & \textbf{Initialization}
& \textbf{Initial} & \textbf{Final} \\
\midrule
Vu et al.
& ACMMP carved & 1.55 & 1.28 \\
Vu et al.
& ACMMP Poisson & 0.61 & 0.62 \\
\textbf{Ours}
& ACMMP carved & 1.55 & \textbf{0.57} \\
\bottomrule
\end{tabular}
\end{table}

\subsection{Photometric bundle adjustment on BMVS}
\label{sec:bundle-adjust}
Since BMVS camera poses may contain calibration errors,
we investigate whether jointly refining camera extrinsics
and surface geometry improves reconstruction accuracy.
We parameterize each camera pose by $\xi_i\in\mathbb R^6$,
so that the camera center $c_i$ and viewing direction
$u_i(p)$ depend on $\xi_i$. For a first-hit point $x=c_i+t_\theta(i,p)u_i(p)$, implicit differentiation gives

\begin{equation}
\frac{\partial x}{\partial \xi_i}
=
\left(I-\frac{u_i\nabla f_\theta(x)^\top}
{\nabla f_\theta(x)^\top u_i}\right)
\left(
\frac{\partial c_i}{\partial \xi_i}
+t_\theta(i,p)\frac{\partial u_i}{\partial \xi_i}
\right).
\end{equation}
\noindent
We minimize the same photoconsistency objective
$\mathcal L_{\mathrm{photo}}(\theta,\{\xi_i\})$, with updated
camera projections and visibility. Starting from the original
camera poses and reconstructed surface, we alternate 500
camera-pose steps (fixing $\theta$, learning rate $10^{-4}$)
and 500 surface steps (fixing $\{\xi_i\}$, learning rate
$10^{-5}$), for 10 cycles. Camera intrinsics remain fixed.
After each pose update, we apply the Umeyama similarity
aligning the current camera centers to their initial positions
to limit global camera-rig drift. BMVS Chamfer distances
are evaluated in the dataset's normalized coordinate frame,
without ground-truth alignment.

\begin{table}[H]
\centering
\caption{Effect of photometric bundle adjustment on BMVS
($\mathrm{CD}_1 \times 10^3$). Per scene: $\mathrm{CD}_1$ (standard deviation of point-to-surface distances) and mean $\pm$ standard deviation over the 8 scenes.}
\label{tab:ba-bmvs}
\small
\setlength{\tabcolsep}{8pt}
\begin{tabular}{lcc}
\toprule
\textbf{Scene} & \textbf{Before BA} & \textbf{After BA} \\
\midrule
Dog        & 1.90\pp{4.34} & \textbf{1.86}\pp{4.34} \\
Bear       & 1.39\pp{2.84} & \textbf{1.20}\pp{2.84} \\
Clock      & 1.46\pp{2.22} & \textbf{1.06}\pp{2.35} \\
Durian     & 2.75\pp{4.23} & \textbf{2.71}\pp{4.22} \\
Man        & 1.57\pp{3.64} & \textbf{1.51}\pp{3.65} \\
Sculpture  & 1.51\pp{2.98} & \textbf{1.46}\pp{2.99} \\
Stone      & 1.72\pp{3.03} & \textbf{1.50}\pp{3.09} \\
Jade       & 3.89\pp{5.05} & \textbf{3.87}\pp{5.05} \\
\midrule
\textbf{Mean} & 2.02\sd{0.87} & \textbf{1.90}\sd{0.94} \\
\bottomrule
\end{tabular}
\end{table}
\subsection{Additional results}
\label{sec:additional-results}
Table~\ref{tab:dtu-per-scene} reports per-scan DTU results, and Fig.~\ref{fig:additional_qualitative} shows additional qualitative comparisons.

\subsection{Visibility}
\begin{figure*}[!p]
    \centering

    \includegraphics[width=\textwidth]{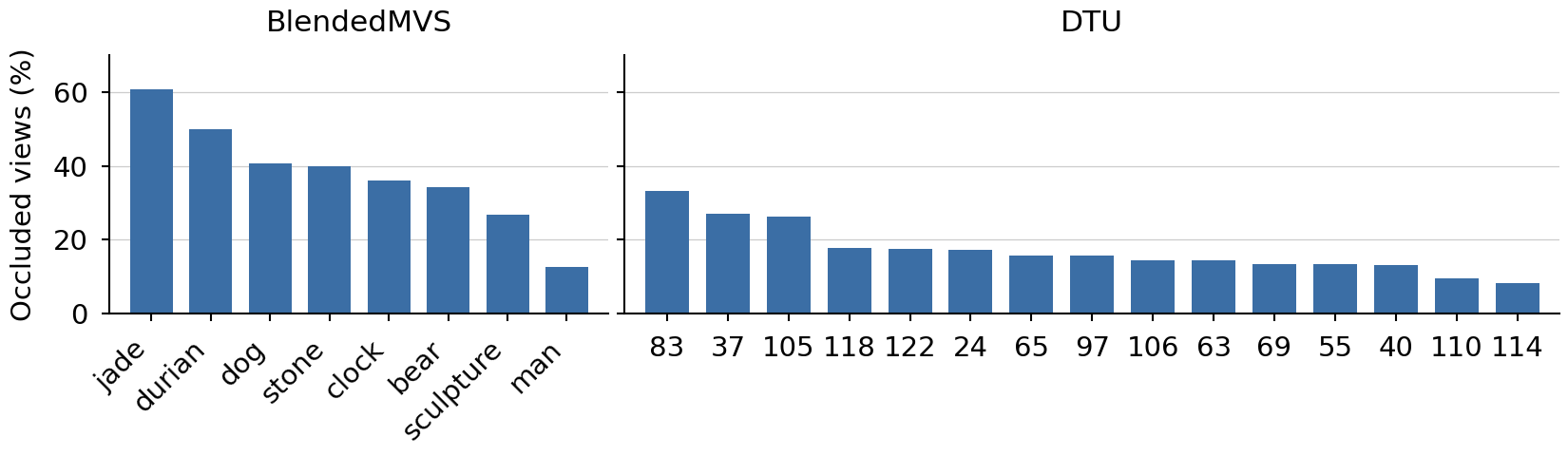}
    \captionof{figure}{Percentage of occluded observations among facing
    views ($n^\top v>0.1$) on BMVS and DTU, estimated from
    60k ground truth surface points per scene.}
    \label{fig:occlusion_prevalence}

    \vspace{12pt}

    \includegraphics[width=0.96\textwidth]{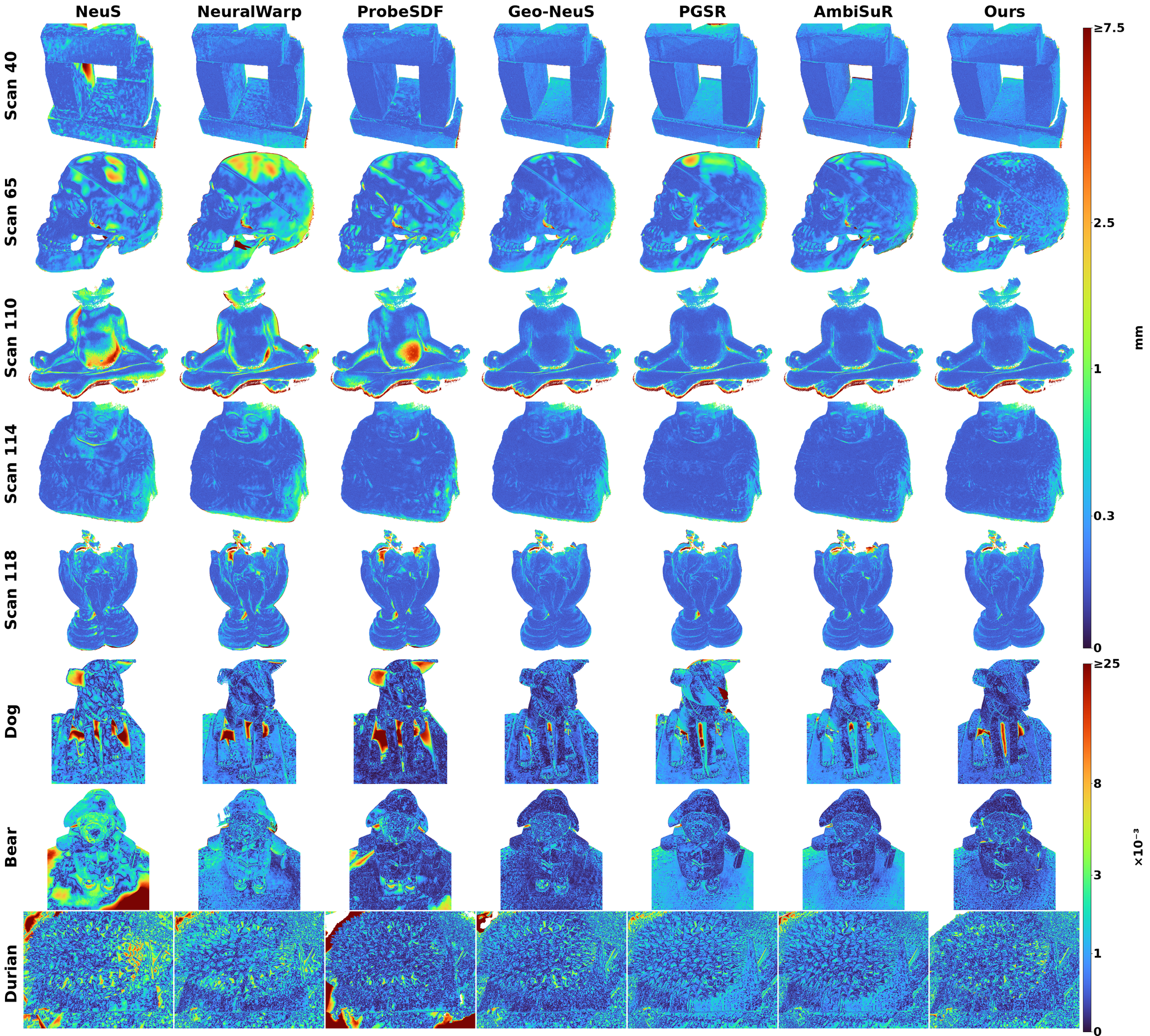}
    \captionof{figure}{Surface reconstruction errors on selected DTU and BMVS scenes.
    Colors indicate distance from the reconstructed surface to the ground truth,
    with separate color scales for DTU (mm) and BMVS ($\times 10^3$ normalized units).}
    \label{fig:additional_qualitative}
\end{figure*}

Fig.~\ref{fig:occlusion_prevalence} reports per-scene occlusion rates and Table~\ref{tab:visibility} summarizes depth localization with and without occlusion filtering.

\begin{table}[H]
\centering
\caption{
Effect of occlusion filtering on DTU depth localization. Mean absolute depth errors (mm) across 15 scans. Filtering has little effect on reliable foreground (0.03 mm reduction, 95$\%$ CI [0.02, 0.05]), but substantially improves partially occluded pixels (0.67 mm reduction, 95$\%$ CI [0.54, 0.81]).
}
\label{tab:visibility}
\small
\begin{tabular}{lccc}
\toprule
& Rays & Without & With \\
\midrule
Reliable foreground & 148k & 1.19 & \textbf{1.15} \\
Partial occlusion & 83k & 2.81 & \textbf{2.13} \\
\bottomrule
\end{tabular}
\end{table}

\begin{figure*}[!p]
\centering

\includegraphics[width=0.86\textwidth]{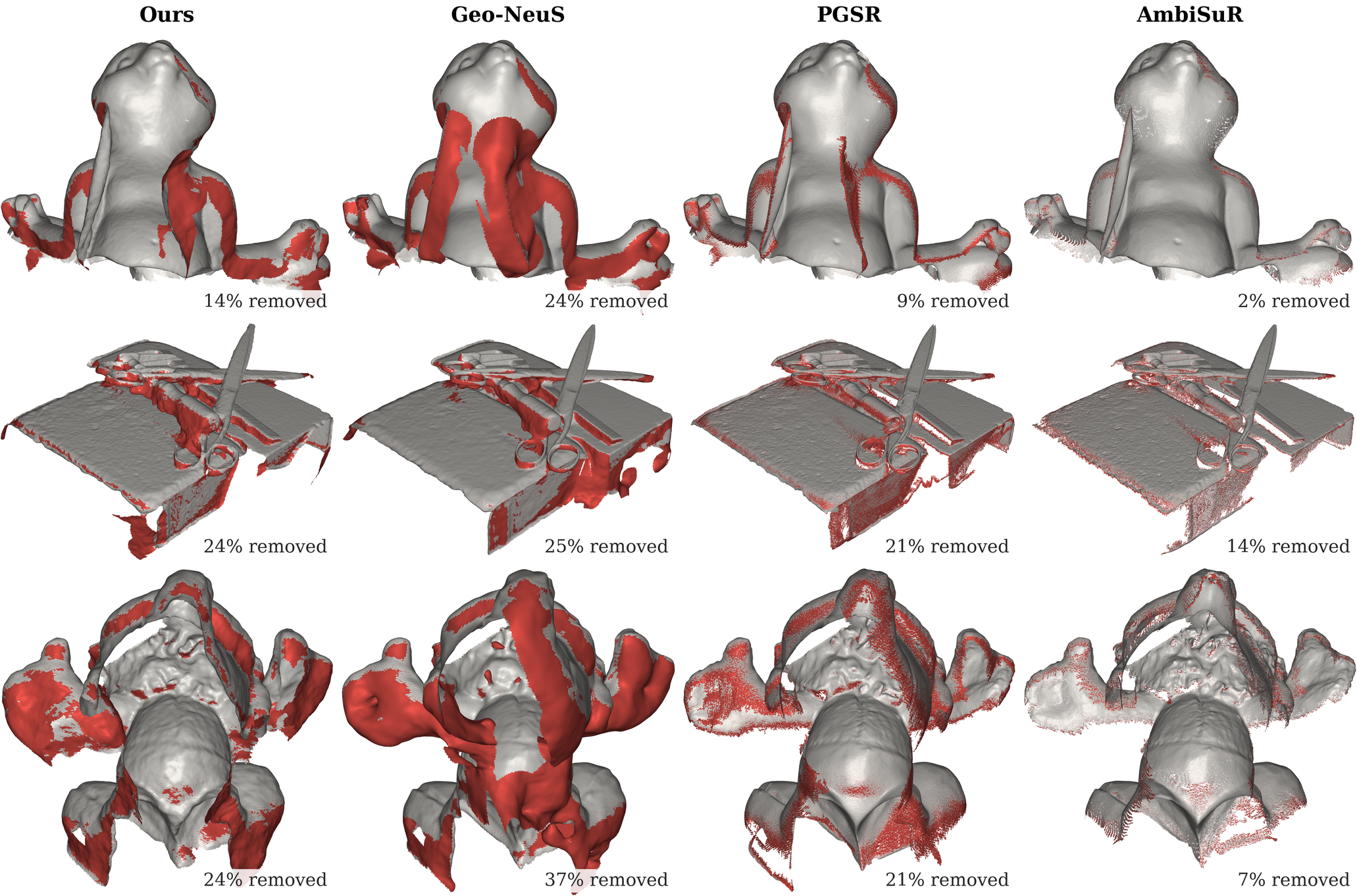}
\caption{Visibility culling on DTU.
Retained surfaces are shown in gray and removed regions in red.
Percentages indicate the fraction of mesh surface area removed.}
\label{fig:culling_visualization}

\vspace{12pt}

\captionof{table}{Extended ablation study on the 15 DTU scans
($\mathrm{CD}_1$ in mm, lower is better). Best results are in bold.}
\label{tab:ablations_extended}
\scriptsize
\setlength{\tabcolsep}{3pt}
\renewcommand{\arraystretch}{1.1}
\resizebox{\textwidth}{!}{%
\begin{tabular}{l*{16}{c}}
\toprule
\textbf{Variant}
& \textbf{24} & \textbf{37} & \textbf{40}
& \textbf{55} & \textbf{63} & \textbf{65}
& \textbf{69} & \textbf{83} & \textbf{97}
& \textbf{105} & \textbf{106} & \textbf{110}
& \textbf{114} & \textbf{118} & \textbf{122}
& \textbf{Mean $\downarrow$} \\
\midrule
No positional encoding
& 1.62 & 1.31 & 0.65 & 0.47 & 0.83
& 1.14 & 1.24 & 1.41 & 1.19 & 0.78
& 0.91 & 0.82 & 0.46 & 1.13 & 0.74
& 0.98\sd{0.34} \\
Single-pixel $\ell_1$
& 0.50 & 0.89 & 0.53 & 0.39 & 1.36
& 1.39 & 0.75 & 1.49 & 1.28 & 0.81
& 0.51 & 1.52 & 0.39 & 0.48 & 0.52
& 0.85\sd{0.43} \\
RGB pixels
& 0.47 & 0.77 & 0.38 & 0.36 & 0.87
& 0.80 & 0.52 & \textbf{1.20} & 0.93 & 0.64
& 0.47 & 0.68 & 0.31 & 0.41 & 0.42
& 0.62\sd{0.26} \\
No occlusion filtering
& 0.42 & 0.76 & 0.37 & \textbf{0.35} & 0.85
& 0.77 & \textbf{0.48} & 1.21 & 0.92 & 0.66
& 0.47 & 0.76 & 0.31 & 0.41 & 0.41
& 0.61\sd{0.26} \\
NeuS MLP + eikonal
& 0.47 & 0.89 & \textbf{0.36} & \textbf{0.35} & 0.86
& \textbf{0.53} & \textbf{0.48} & 1.21 & 0.94 & 0.64
& \textbf{0.46} & 0.66 & \textbf{0.30} & \textbf{0.38} & 0.41
& 0.60\sd{0.27} \\
\midrule
\textbf{Ours}
& \textbf{0.41} & \textbf{0.74} & 0.38 & \textbf{0.35} & \textbf{0.80}
& 0.55 & \textbf{0.48} & \textbf{1.20} & \textbf{0.91} & \textbf{0.62}
& \textbf{0.46} & \textbf{0.62} & \textbf{0.30} & 0.39 & \textbf{0.37}
& \textbf{0.57}\sd{0.25} \\
\bottomrule
\end{tabular}%
}

\vspace{5pt}

\begin{minipage}{0.43\textwidth}
\centering
\includegraphics[width=\linewidth]{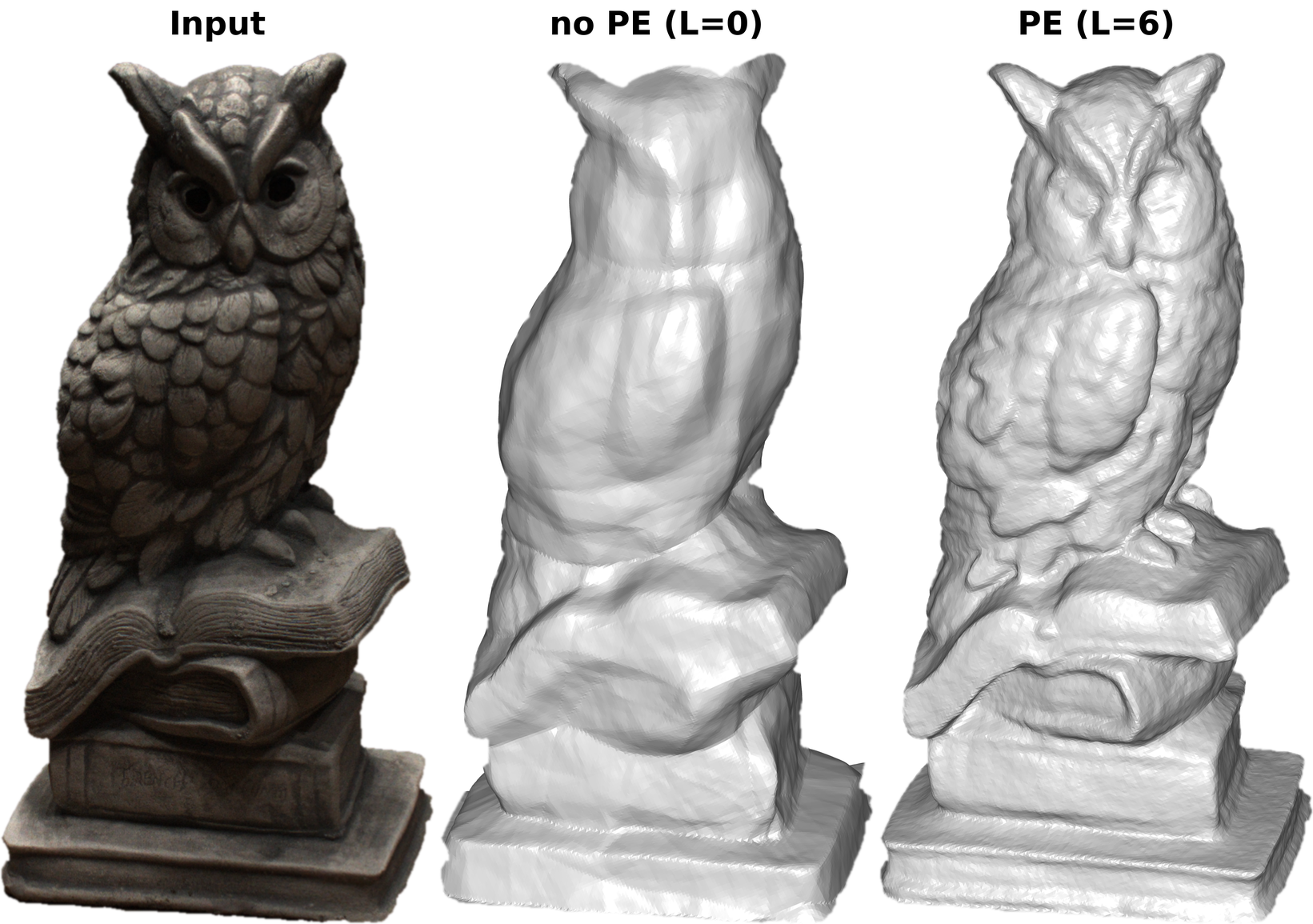}
\captionof{figure}{Effect of positional encoding on DTU scan 122.
Without positional encoding, fine geometric details are lost.}
\label{fig:pe_ablation}
\end{minipage}

\end{figure*}

\begin{table*}[!p]
\centering
\caption{
Surface reconstruction results on DTU ($\mathrm{CD}_1$ in mm).
(A) Published results.
(B) Rerun results with visibility culling; each scan entry gives the Chamfer
distance and, in parentheses, the standard deviation of the point-to-surface
distances in that scan.
Mean column: mean $\pm$ standard deviation over the 15 scans.
Within each panel, the three best results across neural implicit
and Gaussian splatting methods are highlighted in red, orange,
and yellow, respectively.
Rankings use the reported precision, with ties sharing the same rank.
$^\dagger$ indicates the use of foreground masks during optimization.
$^\ast$ For IDR scan 105 in (B), the published unculled result
replaces a failed rerun and is included in the mean.
Best viewed in color.
}
\label{tab:dtu-per-scene}

\scriptsize
\setlength{\tabcolsep}{2.7pt}
\renewcommand{\pp}[1]{\newline{\tiny(#1)}}
\renewcommand{\arraystretch}{1.32}

\resizebox{\textwidth}{!}{%
\begin{tabular}{l*{15}{>{\centering\arraybackslash}p{2.55em}}c}
\toprule
\textbf{Method}
& \textbf{24} & \textbf{37} & \textbf{40} & \textbf{55} & \textbf{63}
& \textbf{65} & \textbf{69} & \textbf{83} & \textbf{97} & \textbf{105}
& \textbf{106} & \textbf{110} & \textbf{114} & \textbf{118} & \textbf{122}
& \textbf{Mean} \\
\midrule

\multicolumn{17}{l}{\textbf{(A) Published results}} \\
\midrule
\multicolumn{17}{l}{\textit{Classical MVS}} \\

PMVS$_{\mathrm{trim}=7}$
& 0.54 & 1.17 & 0.42 & 0.38 & 1.28
& 1.30 & 0.74 & 1.61 & 1.42 & 0.76
& 0.68 & 0.81 & 0.42 & 0.57 & 0.50 & 0.84\sd{0.41} \\

COLMAP$_{\mathrm{trim}=7}$
& 0.53 & 0.99 & 0.38 & 0.36 & 0.99
& 0.85 & 0.52 & 1.37 & 1.03 & 0.60
& 0.59 & 0.59 & 0.29 & 0.42 & 0.42 & 0.66\sd{0.31} \\

ACMMP$_{\mathrm{trim}=7}$
& 0.43 & 0.79 & 0.31 & 0.35 & 0.85
& 0.93 & 0.49 & 1.31 & 0.88 & 0.63
& 0.43 & 0.65 & 0.28 & 0.38 & 0.37 & 0.61\sd{0.29} \\

\midrule
\multicolumn{17}{l}{\textit{Neural implicit surfaces}} \\

IDR$^\dagger$
& 1.63 & 1.87 & 0.63 & 0.48 & 1.04
& 0.79 & 0.77 & 1.33 & 1.16 & 0.76
& 0.67 & 0.90 & 0.42 & 0.51 & 0.53 & 0.90\sd{0.43} \\

NeuS$^\dagger$
& 0.83 & 0.98 & 0.56 & 0.37 & 1.13
& 0.59 & 0.60 & 1.45 & 0.95 & 0.78
& 0.52 & 1.43 & 0.36 & 0.45 & 0.45 & 0.77\sd{0.36} \\

NeuralWarp
& 0.49 & 0.71 & 0.38 & 0.38 & \third{0.79}
& 0.81 & 0.82 & 1.20 & 1.06 & 0.68
& 0.66 & 0.74 & 0.41 & 0.63 & 0.51 & 0.68\sd{0.24} \\

ProbeSDF$^\dagger$
& 0.65 & 0.74 & \second{0.34} & \third{0.34} & 1.02
& 0.71 & 0.62 & 1.34 & 0.94 & 0.70
& 0.53 & 0.96 & 0.36 & 0.45 & 0.47 & 0.68\sd{0.29} \\

Geo-NeuS
& \third{0.38} & \second{0.54} & \second{0.34} & 0.36 & 0.80
& \best{0.45} & \best{0.41} & \second{1.03} & \third{0.84} & \second{0.55}
& \second{0.46} & \second{0.47} & \best{0.29} & \second{0.36} & \third{0.35} & \second{0.51}\sd{0.22} \\

\textbf{Ours}
& 0.41 & 0.74 & 0.38 & 0.35 & 0.80
& \third{0.55} & \third{0.48} & 1.20 & 0.91 & 0.62
& \second{0.46} & 0.62 & \third{0.30} & 0.39 & 0.37 & 0.57\sd{0.25} \\

\midrule
\multicolumn{17}{l}{\textit{Gaussian splatting}} \\

2DGS
& 0.48 & 0.91 & 0.39 & 0.39 & 1.01
& 0.83 & 0.81 & 1.36 & 1.27 & 0.76
& 0.70 & 1.40 & 0.40 & 0.76 & 0.52 & 0.80\sd{0.34} \\

PGSR
& \second{0.36} & \third{0.57} & 0.38 & \best{0.33} & \second{0.78}
& 0.58 & 0.50 & \third{1.08} & \second{0.63} & \third{0.59}
& \second{0.46} & \third{0.54} & \third{0.30} & \third{0.38} & \best{0.34} & \third{0.52}\sd{0.21} \\

AmbiSuR
& \best{0.32} & \best{0.48} & \best{0.31} & \best{0.33} & \best{0.65}
& \second{0.48} & \second{0.42} & \best{0.97} & \best{0.61} & \best{0.54}
& \best{0.36} & \best{0.44} & \best{0.29} & \best{0.34} & \best{0.34} & \best{0.46}\sd{0.18} \\

\midrule
\addlinespace[2pt]
\multicolumn{17}{l}{\textbf{(B) Rerun + visibility culling}} \\
\midrule
\multicolumn{17}{l}{\textit{Classical MVS}} \\

PMVS$_{\mathrm{trim}=7}$
& 0.48\pp{0.78} & 0.98\pp{1.57} & 0.40\pp{0.77} & 0.38\pp{0.72} & 1.15\pp{2.04}
& 1.27\pp{1.99} & 0.70\pp{1.06} & 1.49\pp{2.78} & 1.36\pp{2.28} & 0.72\pp{0.99}
& 0.65\pp{1.37} & 0.77\pp{1.35} & 0.40\pp{0.44} & 0.56\pp{1.00} & 0.46\pp{0.64} & 0.78\sd{0.38} \\

COLMAP$_{\mathrm{trim}=7}$
& 0.41\pp{0.79} & 0.68\pp{1.27} & 0.34\pp{0.76} & 0.36\pp{0.75} & 0.80\pp{1.86}
& 0.74\pp{1.48} & 0.48\pp{0.90} & 1.17\pp{2.55} & 0.96\pp{1.93} & 0.54\pp{0.83}
& 0.50\pp{1.15} & 0.52\pp{1.34} & 0.28\pp{0.32} & 0.38\pp{0.75} & 0.36\pp{0.51} & 0.57\sd{0.26} \\

ACMMP$_{\mathrm{trim}=7}$
& 0.38\pp{0.66} & 0.59\pp{1.24} & 0.30\pp{0.73} & 0.32\pp{0.67} & 0.71\pp{1.86}
& 0.78\pp{1.78} & 0.45\pp{0.90} & 1.22\pp{2.61} & 0.86\pp{1.83} & 0.54\pp{0.89}
& 0.42\pp{0.95} & 0.51\pp{1.12} & 0.28\pp{0.33} & 0.34\pp{0.63} & 0.33\pp{0.52} & 0.54\sd{0.26} \\

\midrule
\multicolumn{17}{l}{\textit{Neural implicit surfaces}} \\

IDR$^\dagger$
& 1.51\pp{2.28} & 1.69\pp{2.89} & 0.59\pp{1.16} & 0.42\pp{0.86} & 0.84\pp{1.83}
& 0.66\pp{1.20} & 0.63\pp{0.91} & 1.12\pp{2.42} & 1.22\pp{2.13} & 0.76$^\ast$\pp{--}
& 0.55\pp{1.17} & 0.74\pp{1.20} & 0.33\pp{0.38} & 0.43\pp{0.77} & 0.41\pp{0.58} & 0.79\sd{0.41} \\

NeuS$^\dagger$
& 0.74\pp{0.96} & 0.83\pp{1.27} & 0.52\pp{1.18} & 0.38\pp{0.78} & 1.01\pp{2.03}
& 0.51\pp{0.88} & 0.51\pp{0.84} & 1.21\pp{2.49} & 1.19\pp{2.15} & 0.78\pp{0.82}
& 0.53\pp{1.11} & 0.85\pp{1.24} & 0.37\pp{0.39} & 0.42\pp{0.76} & 0.40\pp{0.59} & 0.68\sd{0.29} \\

NeuralWarp
& 0.44\pp{0.82} & 0.56\pp{0.88} & 0.34\pp{0.74} & 0.43\pp{0.86} & 0.70\pp{1.50}
& 0.75\pp{1.23} & 0.71\pp{0.90} & \second{1.06}\pp{2.16} & 1.00\pp{1.85} & 0.56\pp{0.74}
& 0.60\pp{1.22} & 0.61\pp{0.97} & 0.37\pp{0.42} & 0.53\pp{0.91} & 0.44\pp{0.60} & 0.61\sd{0.21} \\

ProbeSDF$^\dagger$
& 0.52\pp{1.05} & 0.57\pp{1.04} & \best{0.29}\pp{0.64} & \best{0.32}\pp{0.50} & 0.88\pp{1.97}
& 0.60\pp{1.12} & 0.50\pp{0.87} & 1.20\pp{2.44} & 1.13\pp{2.17} & 0.63\pp{0.77}
& \third{0.47}\pp{1.02} & 0.80\pp{1.33} & 0.32\pp{0.39} & 0.40\pp{0.74} & 0.36\pp{0.50} & 0.60\sd{0.28} \\

Geo-NeuS
& 0.39\pp{0.54} & \second{0.50}\pp{0.97} & 0.33\pp{0.66} & 0.36\pp{0.69} & \third{0.69}\pp{1.72}
& \best{0.45}\pp{0.94} & \best{0.39}\pp{0.67} & 1.08\pp{2.42} & \best{0.83}\pp{1.89} & \best{0.53}\pp{0.75}
& 0.52\pp{1.13} & \third{0.45}\pp{0.91} & 0.30\pp{0.33} & 0.38\pp{0.78} & 0.35\pp{0.50} & \third{0.50}\sd{0.21} \\

\textbf{Ours}
& \third{0.37}\pp{0.63} & 0.52\pp{1.02} & 0.34\pp{0.87} & 0.36\pp{0.72} & \second{0.67}\pp{1.78}
& \second{0.46}\pp{1.01} & \third{0.43}\pp{0.75} & \best{1.04}\pp{2.33} & 0.85\pp{1.75} & \third{0.54}\pp{0.81}
& \best{0.40}\pp{0.84} & 0.48\pp{1.10} & \second{0.29}\pp{0.32} & \best{0.35}\pp{0.64} & \best{0.32}\pp{0.45} & \second{0.49}\sd{0.21} \\

\midrule
\multicolumn{17}{l}{\textit{Gaussian splatting}} \\

2DGS
& 0.45\pp{0.43} & 0.64\pp{1.16} & \third{0.32}\pp{0.65} & 0.41\pp{0.53} & 0.87\pp{2.13}
& 0.82\pp{1.31} & 0.75\pp{1.12} & 1.23\pp{2.55} & 1.14\pp{1.99} & 0.60\pp{0.81}
& 0.68\pp{1.40} & 1.90\pp{2.80} & 0.41\pp{0.42} & 0.65\pp{1.15} & 0.46\pp{0.57} & 0.75\sd{0.41} \\

PGSR
& \second{0.34}\pp{0.62} & \third{0.51}\pp{0.85} & 0.35\pp{0.90} & \second{0.34}\pp{0.78} & 0.72\pp{1.78}
& 0.55\pp{1.02} & 0.46\pp{0.76} & 1.10\pp{2.37} & \best{0.83}\pp{1.74} & 0.55\pp{0.69}
& 0.49\pp{1.21} & \second{0.44}\pp{0.87} & \second{0.29}\pp{0.35} & \third{0.36}\pp{0.68} & \second{0.33}\pp{0.43} & 0.51\sd{0.22} \\

AmbiSuR
& \best{0.33}\pp{0.56} & \best{0.47}\pp{0.78} & \second{0.31}\pp{0.57} & \second{0.34}\pp{0.77} & \best{0.64}\pp{1.62}
& \third{0.48}\pp{0.99} & \second{0.42}\pp{0.66} & \third{1.07}\pp{2.34} & \best{0.83}\pp{1.74} & \best{0.53}\pp{0.68}
& \second{0.41}\pp{0.87} & \best{0.42}\pp{0.87} & \best{0.28}\pp{0.33} & \best{0.35}\pp{0.64} & \second{0.33}\pp{0.48} & \best{0.48}\sd{0.22} \\

\bottomrule
\end{tabular}%
}
\end{table*}

\end{document}